\documentclass[table]{ai2style/ai2}
\usepackage{times}
\usepackage[numbers]{natbib}
\usepackage[utf8]{inputenc} 
\usepackage[T1]{fontenc}    
\usepackage{amsfonts}       
\usepackage{pifont}
\usepackage{microtype}      

\usepackage{graphicx}
\usepackage{capt-of}  
\usepackage{lipsum}   
\usepackage{multirow}%
\usepackage[table]{xcolor}  
\usepackage{xspace}
\usepackage{amsmath}
\usepackage{amssymb}
\usepackage{booktabs}
\usepackage{caption}
\usepackage{subcaption} 
\usepackage{cleveref} 
\usepackage[dvipsnames]{xcolor}
\usepackage{placeins}
\usepackage{gensymb}
\usepackage{pifont}
\usepackage{hyperref}
\usepackage{xcolor}
\usepackage{bold-extra}
\usepackage{url}

\usepackage{tcolorbox} 
\usepackage{zi4} 
\newcommand{\abstractlist}{}

\definecolor{rum_color}{HTML}{7E3DA7}
\definecolor{pi_color}{HTML}{EAB711}

\crefname{figure}{Fig.}{Figs.}      
\Crefname{figure}{Fig.}{Figs.}      

\crefname{table}{Tab.}{Tabs.}       
\Crefname{table}{Tab.}{Tabs.}       
\crefname{section}{Sec.}{Secs.}

\definecolor{maroon}{HTML}{F26035}
\definecolor{yellow}{HTML}{FDBC42}
\definecolor{lavender}{HTML}{734f96}
\definecolor{darkergrey}{HTML}{444444}
\definecolor{midgrey}{HTML}{e6eded}
\definecolor{ai2pink}{HTML}{f0529c}%
\definecolor{ai2midpink}{HTML}{fad3e5}
\definecolor{ai2lightpink}{HTML}{fbecf3}
\definecolor{ai2midwhite}{HTML}{f2e5d9}
\definecolor{ai2offwhite}{HTML}{fbf4ee}
\definecolor{ai2green}{HTML}{0fcb8c}
\definecolor{ai2lightgreen}{HTML}{e7f9f3}
\definecolor{ai2darkgreen}{HTML}{105257}
\definecolor{ai2purple}{HTML}{B932EB}
\definecolor{ai2lightpurple}{HTML}{f7e8fc}
\definecolor{neutralEight}{HTML}{343434}
\definecolor{neutralFive}{HTML}{838383}
\definecolor{neutralThree}{HTML}{bebebe}
\definecolor{neutralOne}{HTML}{dedede}
\definecolor{lightgrey}{HTML}{fafcfc}
\definecolor{maroon}{HTML}{F26035}
\definecolor{yellow}{HTML}{FDBC42}
\definecolor{darkred}{RGB}{156, 39, 33}
\definecolor{darkblue}{RGB}{31, 90, 153}
\definecolor{forestgreen}{rgb}{0.13, 0.55, 0.13}
\definecolor{rum_color}{HTML}{7E3DA7}
\definecolor{pi_color}{HTML}{2F8F4E}
\definecolor{darkgreen}{RGB}{0,100,0}    

\newcommand{\coremark}{\textcolor{ai2pink}{\ding{170}}}
\newcommand{\core}{\textsuperscript{\coremark}}

\def\name{MolmoSpaces\xspace}

\begin{document}

\newif\ifshowcomments
\showcommentsfalse 

\ifshowcomments
\newcommand{\ranjay}[1]{{\color{orange} $[$#1$]^R_K$}}
\newcommand{\dieter}[1]{{\color{violet} [dieter]: #1}}
\newcommand{\maxa}[1]{{\color{darkgreen} $[$#1$]^M_A$}}
\newcommand{\rmh}[1]{{\color{olive} [Rose]: #1}}
\newcommand{\omar}[1]{{\color{blue} [Omar]: #1}}
\newcommand{\MS}[1]{{\color{purple} [Mahi]: #1}}
\newcommand{\yejink}[1]{{\color{red} [Yejin]: #1}}
\newcommand{\winson}[1]{{\color{cyan} [Winson]: #1}}
\newcommand{\jordis}[1]{{\color{brown} [Jordi: #1]}}
\newcommand{\abhayd}[1]{{\color{green} [Abhay: #1]}}
\newcommand{\mayag}[1]{{\color{darkblue} [Maya: #1]}}
\newcommand{\arjung}[1]{{\color{magenta} [Arjun: #1]}}
\newcommand{\snehalj}[1]{{\color{darkred} [Snehal: #1]}}
\newcommand{\wilbertp}[1]{{\color{violet} [Wilbert: #1]}}
\newcommand{\roseh}[1]{{\color{lime} [Rose: #1]}}
\newcommand{\ainaz}[1]{{\color{brown} [Ainaz: #1]}}
\newcommand{\todo}[1]{{\color{pink} [TODO: #1]}}
\else
  \newcommand{\ranjay}[1]{}
  \newcommand{\dieter}[1]{}
  \newcommand{\maxa}[1]{}
  \newcommand{\rmh}[1]{}
  \newcommand{\omar}[1]{}
  \newcommand{\MS}[1]{}
  \newcommand{\yejink}[1]{}
  \newcommand{\winson}[1]{}
  \newcommand{\jordis}[1]{}
  \newcommand{\abhayd}[1]{}
  \newcommand{\mayag}[1]{}
  \newcommand{\arjung}[1]{}
  \newcommand{\snehalj}[1]{}
  \newcommand{\wilbertp}[1]{}
  \newcommand{\roseh}[1]{}
  \newcommand{\ainaz}[1]{}
  \newcommand{\todo}[1]{}
\fi

\newcommand{\rum}{CAP\xspace}
\newcommand{\rumlong}{Contact-Anchored Policies\xspace}
\newcommand{\bench}{MolmoSpaces-Bench\xspace}

\newcommand{\molmoscenes}{MolmoSpaces-Scenes\xspace}

\newcommand{\ithor}{MSCrafted\xspace}
\newcommand{\procthor}{MSProc\xspace}
\newcommand{\procthorobja}{MSProcObja\xspace}
\newcommand{\holodeck}{MSMultiType\xspace}
\newcommand{\digitaltwin}{MSTwin\xspace}

\newcommand{\ithorlong}{MolmoSpaces-Scenes-Crafted\xspace}
\newcommand{\procthorlong}{MolmoSpaces-Scenes-Procedural\xspace}
\newcommand{\procthorobjlong}{MolmoSpaces-Scenes-Procedural-Obja\xspace}
\newcommand{\holodecklong}{MolmoSpaces-Scenes-MultiType\xspace}
\newcommand{\digitaltwinlong}{MolmoSpaces-Scenes-DigitalTwin\xspace}

\newcommand{\graspdataset}{MolmoSpaces-Grasp\xspace}
\newcommand{\objdataset}{MolmoSpaces-Objects\xspace}

\newcommand{\aitwo}{\raisebox{-1.5pt}{\includegraphics[height=1.05em]{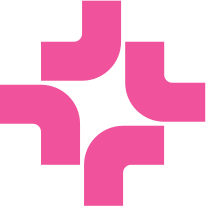}}\xspace}
\newcommand{\hfdataset}{\raisebox{-1.5pt}{\includegraphics[height=1.05em]{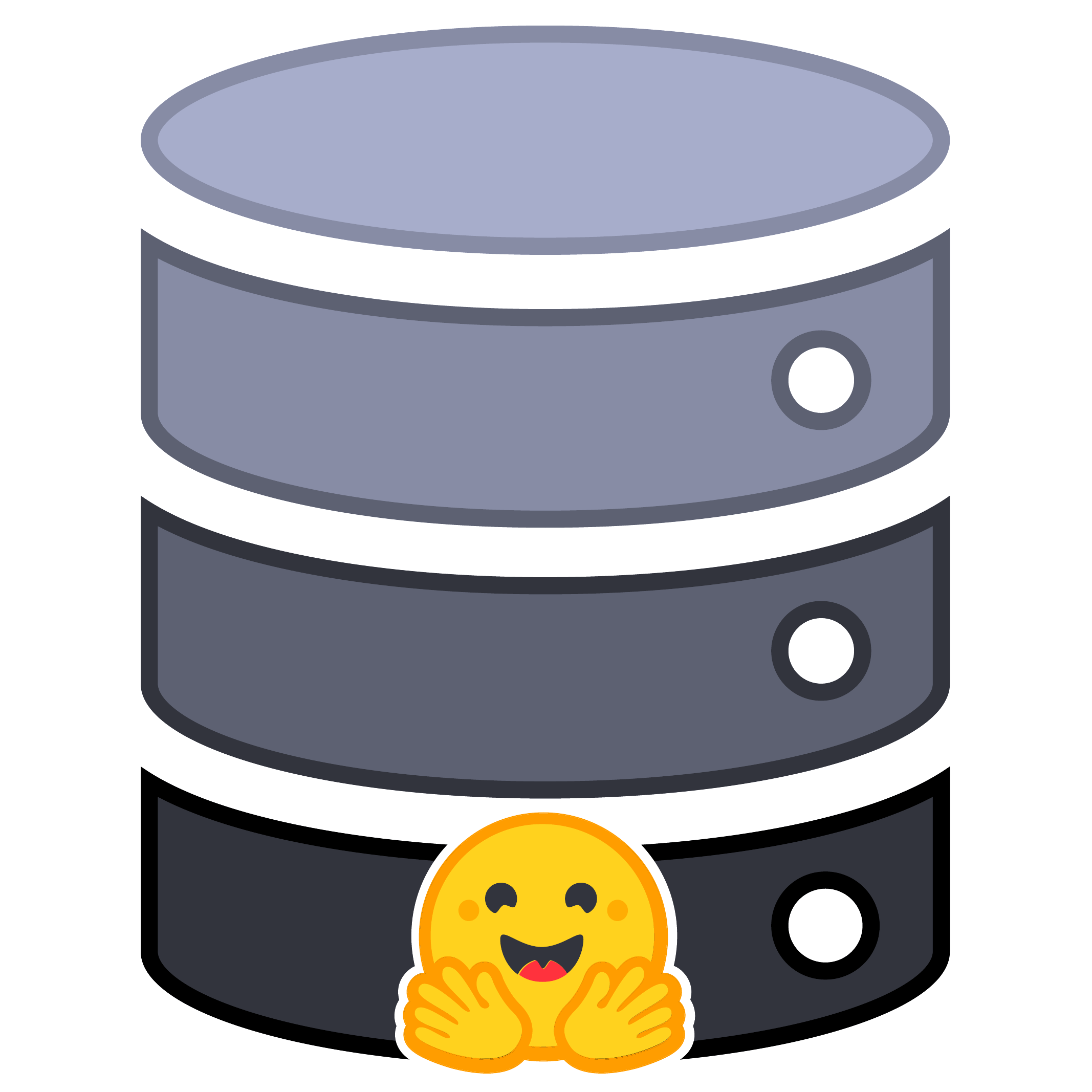}}\xspace}
\newcommand{\github}{\raisebox{-1.5pt}{\includegraphics[height=1.05em]{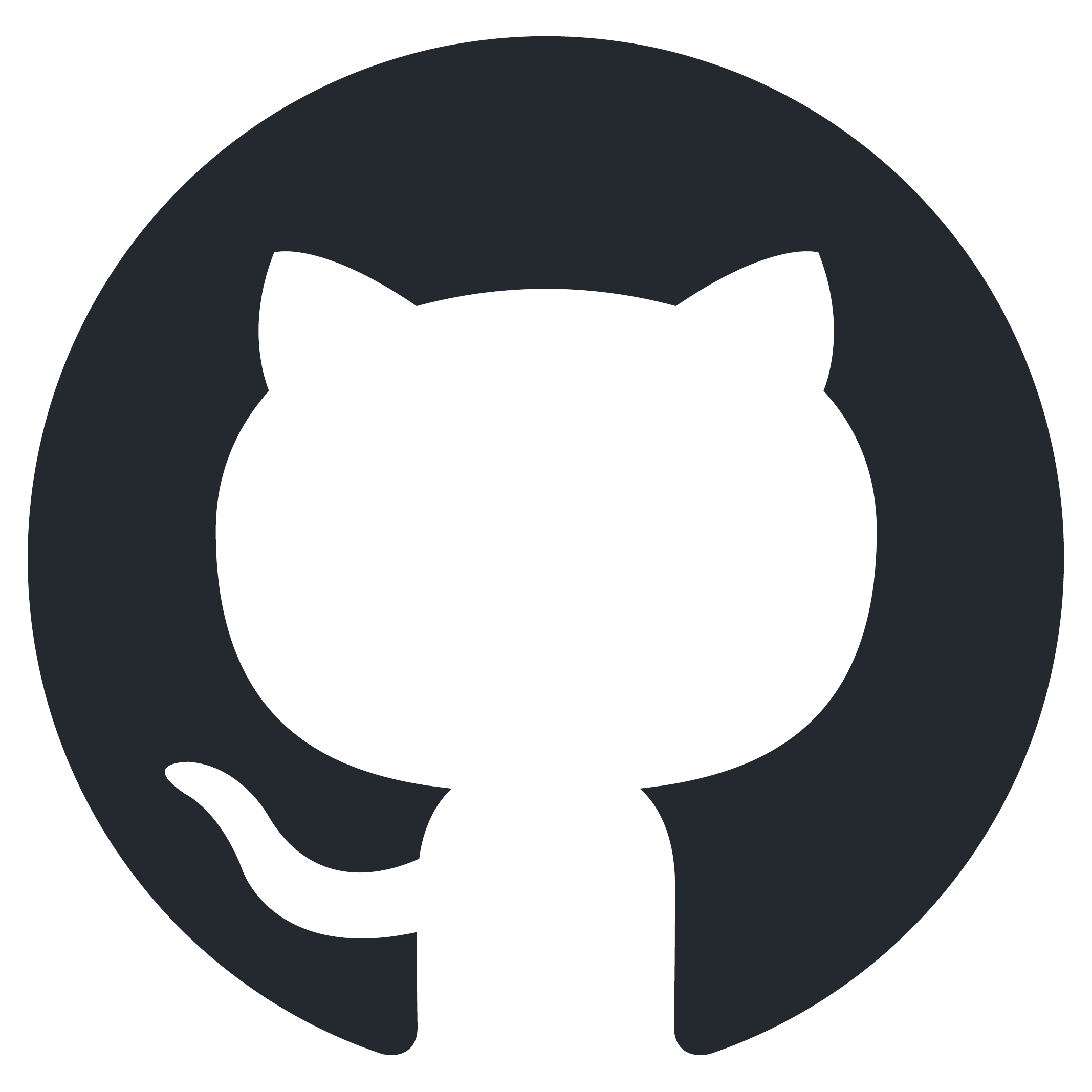}}\xspace}


\title{
\name{\fontsize{15pt}{12pt}\selectfont }\\
{\fontsize{12pt}{12pt}\selectfont A Large-Scale Open Ecosystem for Robot Navigation and Manipulation}
}
\authorOne[1*]{Yejin Kim\core}
\authorOne[1*]{Wilbert Pumacay\core}
\authorOne[3*]{Omar Rayyan\core}
\authorOne[1*]{Max Argus\core}

\authorTwo[1]{Winson Han\core}
\authorTwo[1]{Eli VanderBilt\core}
\authorTwo[1]{Jordi Salvador\core}
\authorTwo[1]{Abhay Deshpande\core}
\authorTwo[1]{Rose Hendrix\core}
\authorTwo[5]{Snehal Jauhri\core}
\authorTwo[2]{Shuo Liu\core}

\authorThree[4]{Nur Muhammad Mahi Shafiullah}
\authorThree[1]{Maya Guru}
\authorThree[2]{Arjun Guru}
\authorThree[2]{Ainaz Eftekhar}
\authorThree[1]{Karen Farley}
\authorThree[2]{Donovan Clay}
\authorThree[1,2]{Jiafei Duan}
\authorThree[1]{Piper Wolters}
\authorThree[1]{Alvaro Herrasti}
\authorThree[2]{Ying-Chun Lee}

\authorThree[5]{Georgia Chalvatzaki}
\authorThree[3]{Yuchen Cui}

\authorFour[1,2]{Ali Farhadi}
\authorFour[1,2]{Dieter Fox}
\authorFour[1,2]{Ranjay Krishna\core}

\affiliation[1]{Allen Institute for AI}
\affiliation[2]{University of Washington}
\affiliation[3]{University of California, Los Angeles}
\affiliation[4]{University of California, Berkeley}
\affiliation[5]{Technische Universität Darmstadt}

\contribution[]{
*denotes equal contribution in no particular order. 
\coremark\ marks core contributors.
See full author contributions \hyperref[sec:contrib]{here}.}

\abstract{
    
Deploying robots at scale demands robustness to the long tail of everyday situations. The countless variations in scene layout, object geometry, and task specifications that characterize real environments are vast and underrepresented in existing robot benchmarks.
Measuring this level of generalization requires infrastructure at a scale and diversity that physical evaluation alone cannot provide.
We introduce \textbf{\name}, a fully open ecosystem to support large-scale benchmarking of robot policies.
\name consists of over 230k diverse indoor environments, ranging from handcrafted household scenes to procedurally generated multiroom houses, populated with 130k richly annotated object assets, including 48k manipulable objects with 42M stable grasps.
Crucially, these environments are simulator-agnostic, supporting popular options such as MuJoCo, Isaac, and ManiSkill. The ecosystem supports the full spectrum of embodied tasks: static and mobile manipulation, navigation, and multiroom long-horizon tasks requiring coordinated perception, planning, and interaction across entire indoor environments. We also design \textbf{\bench}, a benchmark suite of 8 tasks in which robots interact with our diverse scenes and richly annotated objects. Our experiments show \bench exhibits strong sim-to-real correlation ($R$ = 0.96, $\rho$ = 0.98), confirm newer and stronger zero-shot policies outperform earlier versions in our benchmarks, and identify key sensitivities to prompt phrasing, initial joint positions, and camera occlusion. Through \name and its open-source assets and tooling, we provide a foundation for scalable data generation, policy training, and benchmark creation for robot learning research.

}


\metadata[\quad\hfdataset Data:]{
\href{https://huggingface.co/datasets/allenai/molmospaces}{\texttt{Assets and Scenes}}}

\metadata[\quad\github Code:]{
    \href{https://github.com/allenai/molmospaces}{\texttt{https://github.com/allenai/molmospaces}}
}
\metadata[\quad\aitwo Blog:]{\href{https://allenai.org/blog/molmospaces}{\texttt{https://allenai.org/blog/molmospaces}}}

\maketitle

\begin{figure*}[t]
  \centering
  \includegraphics[width=\textwidth]{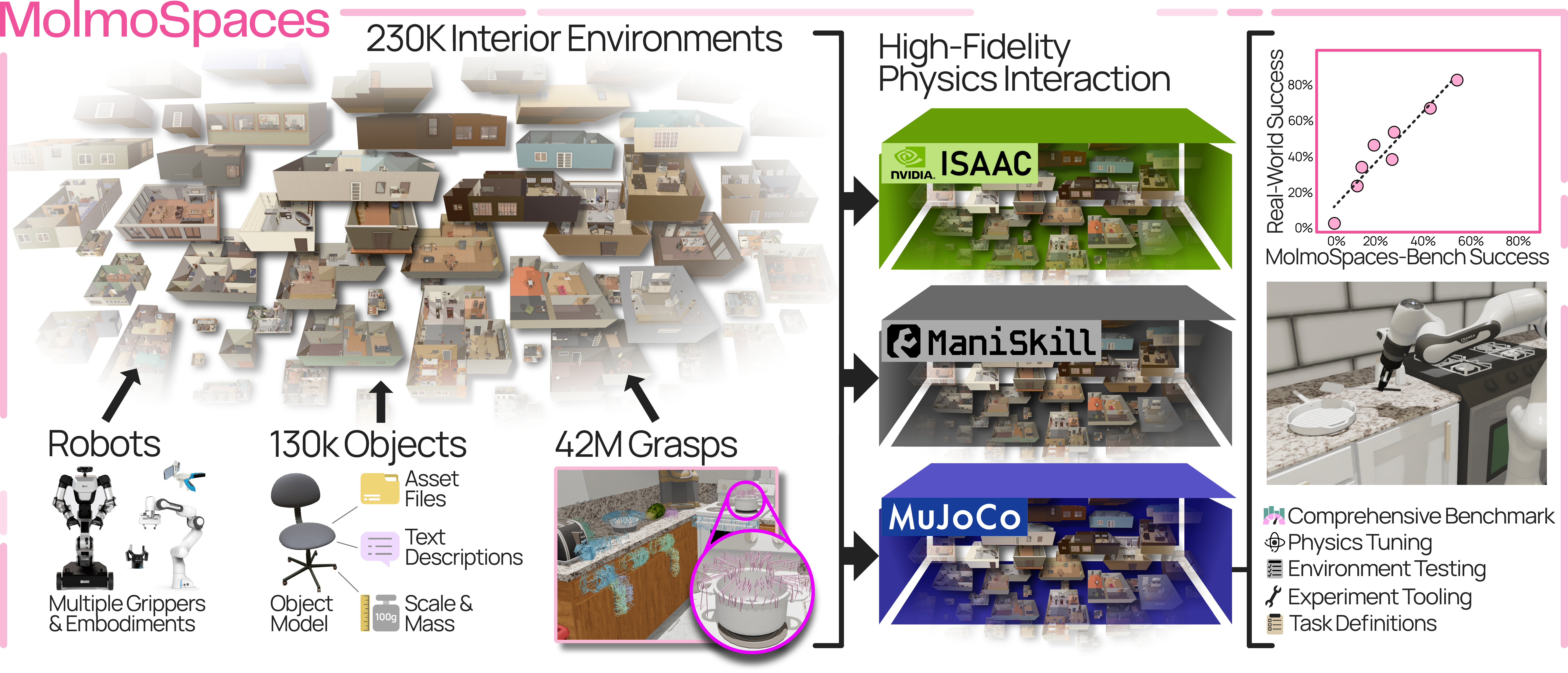}
  \vspace{0.5em}
  \caption{
    \textbf{MolmoSpaces} is an open ecosystem consisting of a large number of
    simulation environments, 3D articulated objects, and tasks for training
    and evaluating robot navigation and manipulation at scale. It provides
    object metadata, grasps, and tooling to generate training data, create
    benchmarks, and evaluate policies in a manner that correlates with
    real-world performance.
  }
  \label{fig:molmospaces_overview}
\end{figure*}


\definecolor{customred}{HTML}{D71B60}   
\definecolor{customteal}{HTML}{40B0A6}  

\newcommand{\ye}{\textcolor{customteal}{\ensuremath{\mathbf{\checkmark}}}}
\newcommand{\no}{\textcolor{customred}{\ensuremath{\mathbf{\times}}}}

\begin{table*}[t]
\centering
\small
\setlength{\tabcolsep}{4pt}
\resizebox{\linewidth}{!}{%
\begin{tabular}{l >{\columncolor{ai2lightpink}}c ccccccccccc}\toprule
\textbf{Feature} &
\textbf{Molmo-} &
\textbf{Robo-} &
\textbf{AI2-} &
\textbf{Habitat} &
\textbf{iGibson} &
\textbf{RL-} &
\textbf{Behavior} &
\textbf{robo-} &
\textbf{Mani-} &
\textbf{OPTIMUS} &
\textbf{LIBERO} &
\textbf{Mimic-} \\
\textbf{} &
\textbf{Spaces} &
\textbf{Casa} &
\textbf{THOR} &
\textbf{2.0} &
\textbf{2.0} &
\textbf{Bench} &
\textbf{1K} &
\textbf{mimic} &
\textbf{Skill 2} &
\textbf{} &
\textbf{} &
\textbf{Gen} \\

\midrule
Scenes                 & 232k & 120   & -- & 1  & 15  & 1   & 50   & 3   & --  & 4   & 20  & 1 \\
Objects                & 130k & 2509  & 3578 & 169 & 1217 & 28 & 9318 & 15 & 2144 & 72 & -- & 40 \\
Object Categories      & 2.8k & 153   & -- & 46 & --  & 28  & 1949 & --  & --  & --  & --   & -- \\
Tasks                  & 8 & 100   & -- & 3  & 6   & 100 & 1000 & 8   & 20  & 10  & 130 & 12 \\
\midrule
Realistic Physics          & \ye & \ye & \no & \no & \ye & \ye & \ye & \ye & \ye & \ye & \ye & \ye \\
Realistic Rendering        &(\ye)& \ye & \ye & \ye & \no & \no & \ye & \no & \ye & \ye & \no & \no \\
Muli-Embodiment            & \ye & \ye & \ye & \no & \ye & \no & \ye & \no & \no & \ye & \no & \ye \\
Room-Scale Scenes          & \ye & \ye & \ye & \ye & \ye & \no & \ye & \no & \no & \no & \no & \no \\
Multi-Room Scenes          & \ye & \no & \ye & \ye & \ye & \no & \ye & \no & \no & \no & \no & \no \\
Annotated Object Grasps          & \ye & \no & \no & \no & \no & \no & \no & \no & \no & \no & \no & \no \\
Mobile Manipulation        & \ye & \ye & \ye & \ye & \ye & \no & \ye & \no & \ye & \no & \no & \ye \\
Scripted Datagen           & \ye & \ye & \no & \no & \no & \ye & \no & \ye & \ye & \ye & \no & \ye \\
AI-generated Tasks         & \ye & \ye & \no & \no & \no & \no & \no & \no & \no & \no & \no & \no \\

\bottomrule
\end{tabular}}
\captionof{table}{Comparison to popular simulation frameworks used in the robot learning literature. The definition of tasks varies strongly across papers; many take it to be unique verb-object or category combinations.}
\label{tab:simulator_comparison}
\end{table*}

\section{Introduction}

Recent advances in robot learning \cite{black2026pi0visionlanguageactionflowmodel,intelligence2025pi05visionlanguageactionmodelopenworld,team2025gemini,bjorck2025gr00t}, have given rise to increasingly general, open-vocabulary policies, capable of zero-shot deployment. As we work towards generalist robots, it becomes important to consider how to evaluate and measure the performance of these policies. 
State-of-the-art models are already nearing saturated performance on several established tasks, providing little signal to drive further progress~\cite{sedlacek2025realm}.
Moreover, most manipulation benchmarks frequently focus on short-horizon skills in a single scene, failing to probe the long-horizon, compositional challenges that arise in realistic environments~\cite{liu2023libero, mees2022calvinbenchmarklanguageconditionedpolicy, james2019rlbenchrobotlearningbenchmark,fei2025liberoplus,zhou2025liberoprorobustfairevaluation,anonymous2026robotarena}.

The real world presents an extraordinarily long tail of situations a robot must handle. Kitchens vary in layout, lighting, and clutter. Objects come in countless shapes, sizes, and materials. Instructions can be phrased in myriad ways. A truly generalist policy must be robust to not just the common cases but to the vast combinatorial space of environments, objects, and tasks that constitute everyday life. Estimating a policy's ability to do so requires evaluating on a far broader distribution of tasks, environments, and objects than ever before.

Simulation offers a compelling path to enable this level of rigor and scale in evaluation. Unlike physical experiments, which are expensive, slow, and difficult to reproduce, simulation enables systematic assessment across thousands of controlled scenarios. 
Rather than testing on a handful of cherry-picked scenarios, we can characterize policy performance across the full distribution of environments a robot might encounter. However, effective simulation for mobile manipulation must simultaneously support scene-scale diversity, physical realism, articulated interactions, and long-horizon compositional tasks in realistic indoor environments. For simulation experiments to be useful, results in simulation must attain strong correlation with real-world performance~\cite{Jain2025PolaRiSSR}. However, existing simulators and benchmarks remain limited. Many provide only dozens of scenes or objects, lack realistic physics or visuals, or support a narrow range of tasks.

We introduce \textbf{MolmoSpaces}, an end-to-end large-scale ecosystem for robotics research illustrated in Figure~\ref{fig:molmospaces_overview}. 
MolmoSpaces unifies diverse scenes, objects, tasks, and tools for training and evaluating generalist robot policies. 
It contains over 230k diverse indoor environments spanning a wide range of layouts and scene types, which enables evaluation across the long tail of real-world spatial configurations. It also includes more than 130k high-quality rigid and articulated object models with rich semantic and physical metadata, which supports assessment of generalization to novel objects. In addition, MolmoSpaces provides over 42M annotated grasps across 48k interactive rigid and articulated objects, which offers ground-truth supervision for grasp success evaluation. Our assets and scenes dataset can be loaded into multiple simulators (MuJoCo~\cite{Todorov2012MuJoCoAP}, IsaacSim~\cite{NVIDIA_Isaac_Sim}, and ManiSkill~\cite{Mu2021ManiSkillGM}), all backed by high-fidelity physics. 


Using this ecosystem, we construct \bench, a new benchmark suite that evaluates robot policies on 8 base tasks--\textit{navigate-to}, \textit{pick}, \textit{pick-and-place}, \textit{pick-and-place-next-to}, \textit{pick-and-place-color}, \textit{open}, \textit{close}, and \textit{open-door} (Sec.~\ref{sec:benchmark_tasks})-- in never-before-seen environments, all zero-shot (i.e. with no fine-tuning on benchmark data). Importantly, the entire MolmoSpaces platform is open-source and extensible. This means that beyond the benchmarks we report, researchers can leverage MolmoSpaces to synthesize their own datasets of scenes, objects, and tasks for training robust robotic policies at scale. We hope that by providing a community-driven ecosystem of this scope, we will accelerate progress toward truly general-purpose robotic intelligence.

In zero-shot evaluations (no task-specific fine-tuning), our benchmark distinguishes performance among several state-of-the-art policies, including VLA models ($\pi$-models \cite{black2026pi0visionlanguageactionflowmodel,pertsch2025fast,intelligence2025pi05visionlanguageactionmodelopenworld}) and classical modular baselines, across a wide range of unseen environments and objects. The results reveal steady progress over model generations, but also expose brittleness to distribution shifts. For instance, we find that minor changes in instruction phrasing or initial robot pose can cause significant drops in success for some policies, especially earlier-generation VLAs. This sensitivity highlights the importance of training on more diverse data, which MolmoSpaces can readily supply for future work. 
Encouragingly, we observe a strong sim-to-real correlation: policies that score higher in our simulation benchmarks also achieve better real-world success rates on equivalent tasks (with Pearson $R^2 \approx 0.92$ for object picking). This validates that high-fidelity simulation can be a reliable proxy for real-world performance. Moreover, by systematically perturbing scene parameters and sensor inputs in simulation, we pinpoint specific failure modes (e.g. dependence on certain camera viewpoints and lighting conditions) that are costly to uncover with physical trials. These analyses demonstrate how an open ecosystem like MolmoSpaces not only measures overall progress but also yields insights to drive algorithm improvement.

By dramatically expanding the scale of available simulated environments and making them openly accessible, MolmoSpaces enables researchers to measure generalization more rigorously than ever before and can support future work in generating diverse training data for tackling the next generation of robotics problems. 

\section{Related work}

\textbf{Robot simulation frameworks} provide scalable, safe, and repeatable platforms for rapid prototyping, policy learning, and evaluation. Modern simulators such as MuJoCo, Isaac, and ManiSkill offer high-fidelity physics simulations that support contact- and force-based manipulation~\cite{Todorov2012MuJoCoAP,NVIDIA_Isaac_Sim,Mu2021ManiSkillGM,Yin2026GenieS3}. Building on these engines, the research community has developed a range of simulation frameworks.
Projects like AI2-THOR~\cite{kolve2017ai2thor}, Habitat, and Habitat-Lab~\cite{savva2019habitat, szot2022habitat20traininghome} emphasize photorealistic visual navigation, but provide only limited manipulation support, often relying on “magic grasps” that bypass realistic contact dynamics~\cite{kolve2017ai2thor, deitke2022procthor}. RoboCasa and RoboCasa365~\cite{robocasa2024, anonymous2026robocasa} build on MuJoCo to provide single-room environments, synthetic datasets, and task definitions for multi-task manipulation and navigation, but remain limited in scene and asset diversity.

\textbf{Large-scale datasets in robotics} are increasingly important, as demonstrated by efforts such as Open X-Embodiment~\cite{o2024open}. In parallel, the community has pursued data scaling through simulation. Objaverse~\cite{deitke2023objaverse} provides internet-scale 3D assets compatible with simulators, while ProcTHOR expands AI2-THOR with tens of thousands of procedurally generated multi-room houses~\cite{deitke2022procthor}. Holodeck~\cite{yang2024holodeck} introduces LLM-guided scene generation beyond household environments, and InternScenes~\cite{zhong2025internsceneslargescalesimulatableindoor} combines real scans, procedural layouts, and designer-created environments to provide diverse indoor scenes at scale. Despite addressing scene and asset scale, these datasets offer limited support for physics-based manipulation and are primarily evaluated on navigation tasks. Conversely, GraspGen~\cite{Murali2025GraspGenAD} provides large-scale grasp annotations for Objaverse assets, but these remain at the asset level and require substantial effort to integrate into interactive scenes.

By contrast, \textbf{\name} addresses both scale and task diversity through a ready-to-use ecosystem that unifies 230K scenes from AI2-THOR~\cite{kolve2017ai2thor}, ProcTHOR~\cite{deitke2022procthor}, and Holodeck~\cite{yang2024holodeck}, and makes them compatible across MuJoCo, IsaacSim, and ManiSkill. The ecosystem further incorporates 130K object assets from Objaverse~\cite{deitke2023objaverse}, with over 48K objects annotated with grasp data and validated to be pickable and articulable under realistic physics. Together, these components enable scalable, diverse, and physically grounded evaluation of navigation, manipulation, and mobile manipulation policies. Comparisons between \name and prior work are summarized in Table~\ref{tab:simulator_comparison}.

\medskip
Benchmarks are central to progress in robotics, providing standardized tasks and evaluation protocols for fair comparison, reproducibility, and diagnosis of failure modes. However, real-world benchmarking remains difficult to scale due to hardware heterogeneity, differences in sensing and control stacks, and the time and labor required for evaluation. Recent efforts such as RobotArena~\cite{Atreya2025RoboArenaDR} partially address these challenges through distributed, crowd-sourced evaluation, while others like AutoEval~\cite{zhou2025autoeval} leverage success classifiers and reset policies to facilitate near-autonomous real-world evaluations of specific tasks. However, real-world evaluation alone remains limited in the scale and diversity needed to robustly assess generalist policies.

\textbf{Simulation-based benchmarks} offer a scalable and reproducible alternative, enabling controlled variation and systematic stress testing that are impractical in the real world. A wide range of benchmarks have emerged for manipulation~\cite{liu2023libero,james2019rlbenchrobotlearningbenchmark} and navigation~\cite{kolve2017ai2thor,shridhar2020alfredbenchmarkinterpretinggrounded,habitatchallenge2023}, becoming standard testbeds for evaluation. RoboVerse~\cite{Geng2025RoboVerseTA} unifies several of these benchmarks under a shared framework. As policies converge on vision–language–action (VLA) models, recent benchmarks such as LIBERO~\cite{liu2023libero}, CALVIN~\cite{mees2022calvinbenchmarklanguageconditionedpolicy}, LIBERO-Plus~\cite{fei2025liberoplus}, LIBERO-Pro~\cite{zhou2025liberoprorobustfairevaluation}, VLABench~\cite{zhang2024vlabench}, and RobotArena-Infinity~\cite{anonymous2026robotarena} expand evaluation to include language grounding and generalization. Additional works like the COLOSSEUM~\cite{pumacay2024colosseum} and VLATest~\cite{wang2025vlatest} evaluate the robustness of generalist policies to changes in lighting, distractor objects, camera poses, and more. Other efforts, including BEHAVIOR-1K~\cite{li2022behavior}, ManiSkill-HAB~\cite{Shukla2024ManiSkillHABAB}, and EmbodiedBench~\cite{yang2025embodiedbenchcomprehensivebenchmarkingmultimodal}, extend benchmarks to mobile manipulation, but remain biased toward household environments and limited in scene scale.

\textbf{Sim-to-real benchmarks} focus on providing simulation evaluations that match real-world results. SIMPLER~\cite{li24simpler} studies distributional shifts by pairing simulated evaluations with real-robot rollouts, while PolaRiS~\cite{Jain2025PolaRiSSR} constructs digital twins from real-world videos and demonstrates strong sim-to-real correlation across realistic scenes, albeit limited to tabletop manipulation and partial environment reconstruction. Other works analyze large behavior models across simulated and real environments, but rely on proprietary evaluation pipelines and closed datasets. Moreover, many sim-to-real benchmarks require training on simulation data, limiting their ability to assess true zero-shot generalization~\cite{Jain2025PolaRiSSR,li2022behavior}.

In contrast, our benchmark evaluates zero-shot generalist policies across navigation, manipulation, and mobile manipulation tasks using a validation set of over 20K diverse indoor scenes—spanning both household and non-household environments—and more than 22K interactable rigid and articulated objects. It supports large-scale distributional analysis under controlled perturbations to scenes, objects, sensors, and language prompts, enabling rigorous assessment of generalization and failure modes.

\section{MolmoSpaces}

MolmoSpaces contains:
\begin{enumerate}
\item \textbf{\molmoscenes:} Over 230k diverse indoor environments spanning a wide range of layouts and scene types, enabling evaluation across the long tail of real-world spatial configurations.
\item \textbf{\objdataset:} More than 130k high-quality rigid and articulated object models with rich semantic and physical metadata, supporting assessment of generalization to novel objects.
\item \textbf{\graspdataset:} Over 42M annotated grasps across 48k interactive rigid and articulated objects, providing ground-truth supervision for grasp success evaluation.
\item \textbf{\bench:} A benchmark suite comprising object-centric tasks across 8 task types, zero-shot evaluations with no fine-tuning, and sim-to-real correlation analysis demonstrating that lessons learned in simulation transfer to real-world performance.
\item \textbf{Simulation Infrastructure:} Scalable tooling for task composition, benchmark creation\yejink{undecided to leave this or not. This is mentioned in Introduction}, and reproducible evaluation. 
\end{enumerate}

\subsection{\molmoscenes}
\label{sec:scene_datasets}

\begin{figure}[t]
    \centering
    \includegraphics[width=\textwidth]{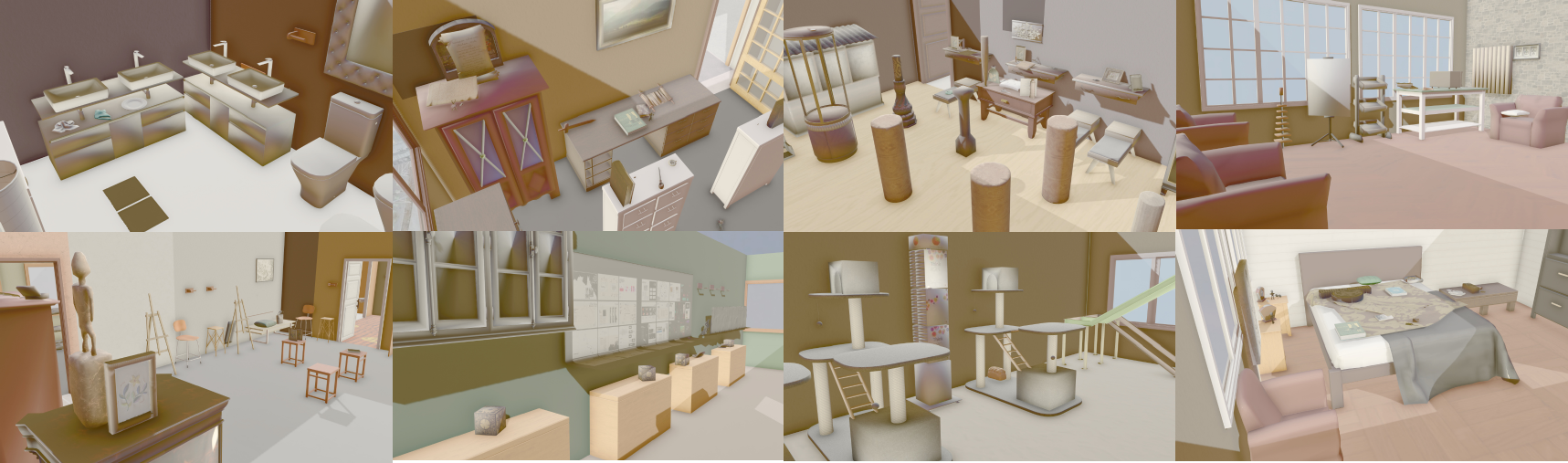} 
    \caption{Examples of diverse scene environments from \holodecklong with the Filament rendering engine. Our ecosystem contains scenes from art studies, cat cafes, lounges, museums, and many other scenes, all pre-populated with objects to be manipulated.}
    \label{fig:diverse_environment}
\end{figure}

We provide five scene datasets, summarized in Table~\ref{tab:scene-datasets-transposed}. The scenes were originally sourced from multiple datasets in AI2-THOR: \ithorlong, \holodecklong, \procthorlong, and \procthorobjlong. We process and tune these scenes to be physically realistic in MuJoCo, ManiSkill, and IsaacSim. \ithorlong (\ithor) contains 120 hand-crafted single-room scenes evenly distributed among kitchens, bedrooms, living rooms, and bathrooms, split into 48/48/24 train/validation/test scenes, with object placement carefully curated to ensure physical stability. \procthorlong (\procthor) provides 12k procedurally generated residential scenes with 10k/1k/1k train/validation/test splits, where each scene represents a house with between one and ten rooms with layouts designed for realistic household navigation and manipulation. \procthorobjlong (\procthorobja) extends this with 110k train/validation scenes containing both THOR and Objaverse objects. \holodecklong (\holodeck) provides 110k diverse scene types generated via LLM-based procedural generation, also with THOR and Objaverse objects. Finally, \digitaltwinlong (\digitaltwin) is a high-fidelity manual reconstruction of our real-world kitchen. All datasets contain both rigid and articulated objects. Some examples of these scenes are shown in Figure~\ref{fig:diverse_environment}.

\begin{table*}[t]
\centering
\small
\begin{minipage}[t]{0.48\textwidth}
    \vspace{15pt} 
    \centering
    \renewcommand{\arraystretch}{1.2} 
    \resizebox{\linewidth}{!}{%
    \begin{tabular}{lcccc}
    \toprule
    \textbf{Dataset} & \textbf{Scene} & \textbf{Object} & \textbf{Obj /} & \textbf{Creation} \\
    \textbf{Name} & \textbf{Count} & \textbf{Set} & \textbf{Scene} & \textbf{Method} \\
    \midrule
    \digitaltwin  & 1 & - & 8 & Manual \\
    \ithor & 120 & THOR & $\sim$30 & Hand-crafted \\
    \procthor & 12k & THOR & $\sim$72 & Heuristic \\
    \procthorobja & 110k & THOR+ & $\sim$60 & Heuristic \\
    \holodeck & 110k & THOR+ & $\sim$97 & LLM Proc. \\
    \bottomrule
    \end{tabular}}
    \vspace{2pt} 
    \caption{\molmoscenes contains five scene datasets with varying scene types, scales, objects, and creation methods.}
    \label{tab:scene-datasets-transposed}
\end{minipage}
\hfill
\begin{minipage}[t]{0.48\textwidth}
    \vspace{0pt} 
    \centering
    \scriptsize
    \renewcommand{\arraystretch}{1.0}
    \resizebox{\linewidth}{!}{%
    \begin{tabular}{lcccc}
    \toprule
    \textbf{Dataset Name} & Stab. & Lift & Inter. & Artic. \\
    \midrule
    \ithor & 100.0 & 97.5 & 98.3 & 99.1 \\
    \procthor & 99.4 & 99.7 & 99.9 & 97.4 \\
    \procthorobja & 98.4 & 98.7 & 99.1 & 93.7 \\
    \holodeck & 94.9 & 99.6 & 93.7 & 65.2 \\
    \bottomrule
    \end{tabular}}
    \caption{Scene datasets are all quality tested in the MuJoCo simulator with high pass rates.}
    \label{tab:scene-testing}
    
    \vspace{1.2em} 

    \centering
    \resizebox{\linewidth}{!}{%
    \begin{tabular}{lcccc}
    \toprule
    \textbf{Dataset Name} & MSCraft & MSProc & MSObja & MSMulti \\
    \midrule
    Stability test (\%) & 92.5 & 93.9 & 95.9 & 95.2 \\
    \bottomrule
    \end{tabular}}
    \caption{Isaac Sim simulator pass rates.}
    \label{tab:scene-isaac-testing}
\end{minipage}
\end{table*}

To generate \holodeck, we extend the diverse scene generation pipline presented in Holodeck \cite{yang2024holodeck}. We select and extend indoor scene types in the SUN database~\cite{xiao2010sun}, based on the suitability to the available THOR and Objaverse object taxonomy, and reorganize them into a hierarchy with between five and fifteen concrete scene types for each of ten generic types. In total, 546 room types (including many scene-specific) and 101 scene types are available for scene sampling (Fig.~\ref{fig:holodeck_objaverse_distribution} left, in the appendix). Each scene specification contains a generic or concrete scene type and between one and ten rooms of diverse types. We also sample from a subset of 52k persona descriptions from~\cite{ge2024scaling} -- chosen by their suitability to produce visual and stylistic differences in objects, materials, or layout constraint selection -- and accentuate some particular style in 90\% of the scene specifications, which are finally converted to text prompts for LLM scene generation. We add a ‘grid’ constraint to the DFS-based object placement optimizer in Holodeck to simplify the uniform placement of objects in available free space commonly occurring in non-residential scenes. 

\begin{figure}[t]
    \centering
    \includegraphics[width=\textwidth]{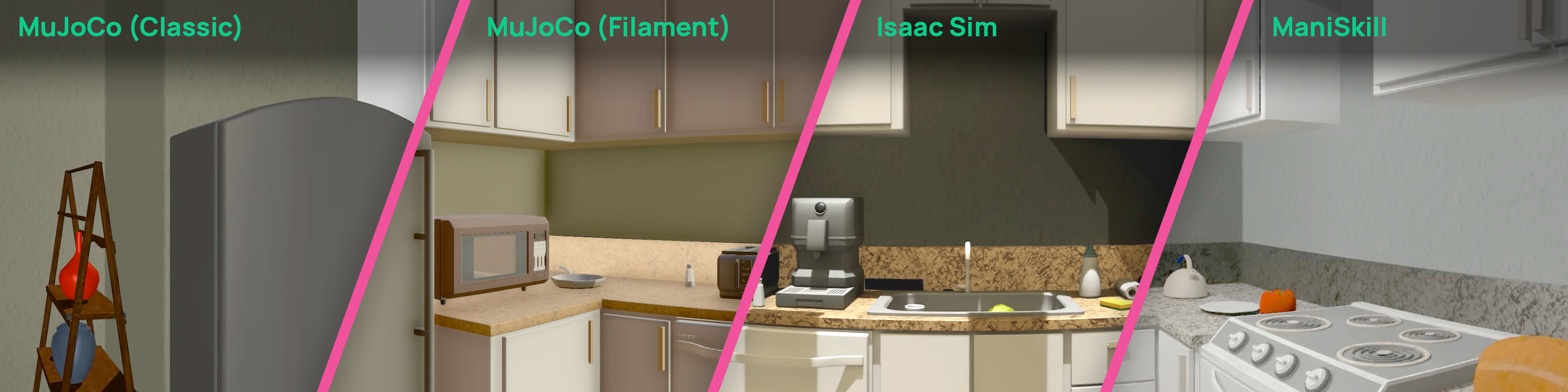} 
    \caption{An example scene rendered across different simulators: MuJoCo, Issac Sim, and ManiSkill. When using MuJoCo, the scenes can be rendered using either the OpenGL renderer (Classic) or with Filament (Filament).}
    \label{fig:multi_simulator_pan}
\end{figure}

In order to enable the wide-spread use of our simulation assets, we also release them converted to the USD format, which can be natively loaded in IsaacSim. Additionally, we provide both an asset and scene loader for ManiSkill. Figure~\ref{fig:multi_simulator_pan} illustrates the same scene rendered across aforementioned simulators and their respective rendering engines. Finally, we generate occupancy maps for all scenes to identify collision-free starting poses for robots, ensuring safe initialization for both manipulation and navigation experiments.

\textbf{Scene quality testing:} For an environment to be suitable for both navigation and manipulation tasks, objects must remain physically stable, respond appropriately to applied forces, and be accessible for interaction. Specifically, objects should not drift or randomly move, rigid objects must be pick-upable, articulated objects must allow motion across their defined ranges, and masses, friction, and other physical parameters should yield realistic dynamics under standard forces. Ensuring these properties is particularly important when integrating assets not originally designed for physics-based
simulation. Additionally, objects that are originally designed for such simulations often have innately incompatible parameters by default which also warrant additional tuning for integration.

To systematically enforce these conditions, we implemented four tests addressing different aspects of physical validity: scene stability, object intersections, rigid object liftability, and articulation of movable objects. First, each scene was settled by simulating approximately 20 seconds and then saved with the updated, settled poses. Stability tests simulate the environment for multiple steps after settling and remove any objects that continue to jitter. Intersection tests detect colliding objects: if a collision occurs between a fixed and a free object, the free object is removed; if the collision is between two free objects, the smaller one is removed. Lift tests apply upward forces to all free objects and measure their displacement along the z-axis; objects that cannot be lifted by at least 5 cm and are detected inside another object’s site are removed. Articulation tests apply forces at joints to open or close articulated objects; if an object cannot be actuated through at least 70\% of its joint range, free objects located within its articulation site or blocking its motion are removed

Across MuJoCo scenes (Table~\ref{tab:scene-testing}), over 95\% of environments pass these stability and manipulation checks. The lowest success rate is observed for articulation tests on the MSMultiType scene dataset (63\%). We note that these scenes were designed to maximize navigability by biasing large object placement to prevent blocking door-to-door navigation while ensuring many small objects can be accessibly placed for manipulation tasks. Manual inspection confirms that most failures arise from scene layout rather than limitations of the simulation engine. We additionally evaluated a subset of scenes in Isaac Sim, observing consistently high pass rates as shown in Table~\ref{tab:scene-isaac-testing}, which demonstrates the robustness and cross-platform validity of our validation pipeline.


\subsection{\objdataset}

\begin{table*}[]
    \centering
    \begin{tabular}{lllllll}
    \toprule
    Dataset & {Total} & {Pickupable} & {Non-Pick} & {Articulated} & THOR & Objaverse \\
    \midrule
    MSCrafted   & ${\sim}30$  & ${\sim}14$ & ${\sim}7$  & ${\sim}9$  & ${\sim}23$  &     --      \\
    MSProc      & ${\sim}72$  & ${\sim}34$ & ${\sim}31$ & ${\sim}7$  & ${\sim}41$  &     --      \\
    MSProcObja  & ${\sim}105$ & ${\sim}47$ & ${\sim}50$ & ${\sim}7$  & ${\sim}72$  & ${\sim}32$  \\
    MSMultiType & ${\sim}150$ & ${\sim}61$ & ${\sim}77$ & ${\sim}12$ & ${\sim}73$  & ${\sim}77$   \\
    \bottomrule
    \end{tabular}
        \caption{Average number of objects per scene on each dataset}
    \label{tab:object_statistics}
\end{table*}

\begin{figure}[t]
    \centering
    \includegraphics[width=\textwidth]{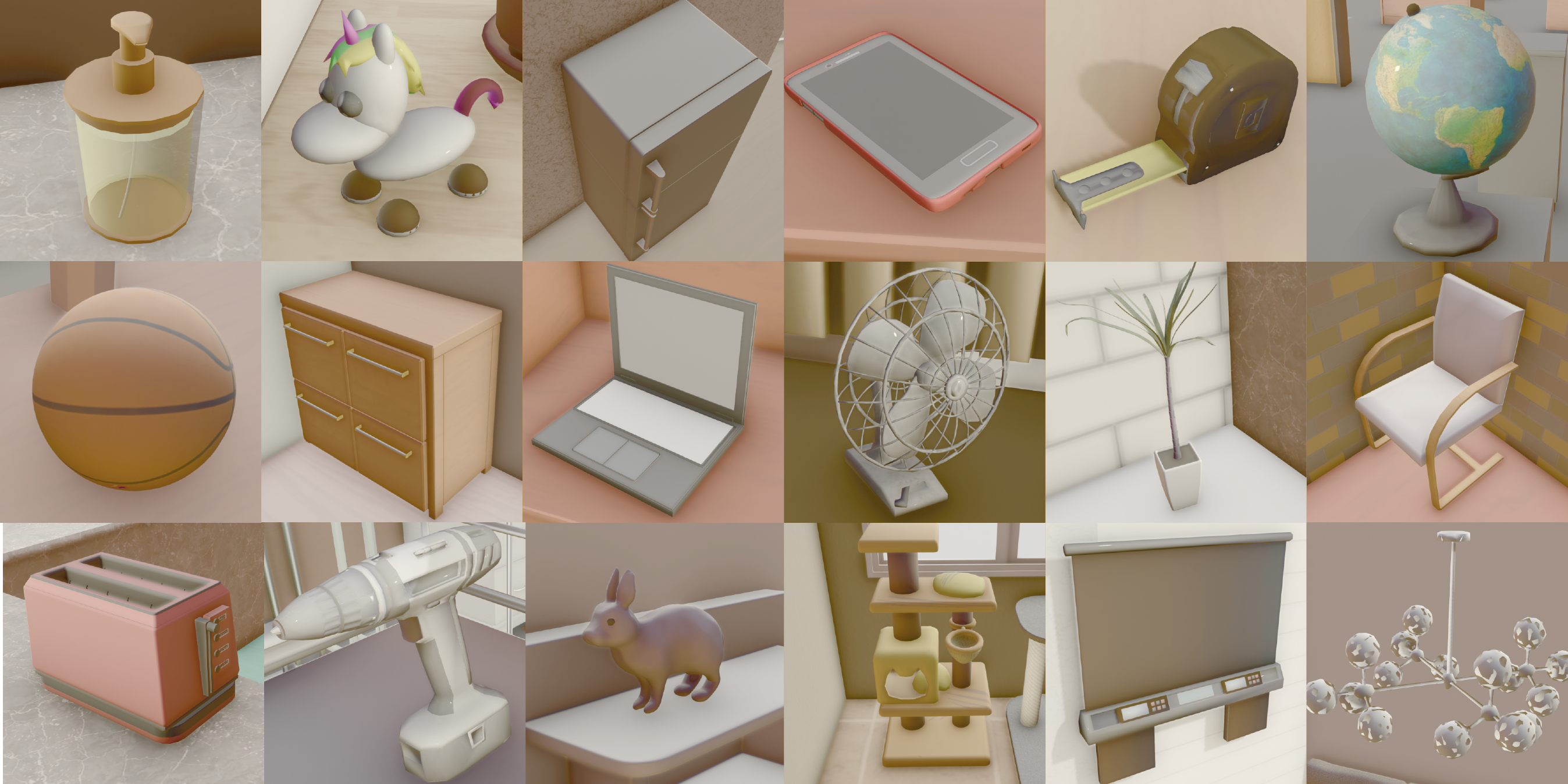} 
    \caption{A random sampling of object types in our ecosystem, with different sizes, shapes, and articulations. These examples are rendered with Filament.}
    \label{fig:object_images}
\end{figure}

We provide two object model datasets consisting of 1.6k THOR and 129k Objaverse objects, with samples shown in Figure~\ref{fig:object_images}. These assets populate the generated scenes described in Section~\ref{sec:scene_datasets}. Table~\ref{tab:object_statistics} reports the average number of objects per scene for each dataset. To ensure physical realism, rigid objects were validated by estimating mass and density against LLM-estimated values, and articulable objects were tuned by manipulating them via teleoperation of a simulated Franka FR3 robot. 
%
Collider meshes were generated using COACD~\cite{wei2022coacd}, with primitive colliders \textit{human annotated} for all THOR assets. For stability, receptacle objects primarily use primitive colliders, while manipulable objects use convex decomposition except for very small or thin items, where primitives are preferred. Meshes in Objaverse models were further processed and decimated for simulation efficiency \cite{ai22024objathor}.

THOR objects span 134 categories, with 22 articulable categories (e.g., doors, refrigerators) annotated with joint types, axes, positions, and ranges.
Objaverse objects, spanning almost 2.8k WordNet~\cite{mccrae2019wordnet,mccrae2022wordnet} synsets, are curated from 625k models annotated with descriptions, mass estimates, canonical poses, pickable and receptacle properties, synsets, and scale estimates generated by GPT-4o~\cite{achiam2023gpt,hurst2024gpt}.
A complementary GPT-4.1~\cite{openai_gpt41_api_2024} annotation identified object counts, extraneous or missing geometry, texture quality, and receptacle presence in models with parseable annotation.
Filtering provided 129k single-object models that met scale consistency, sufficient texture quality, cross-renderer fidelity, compact file size, collider quality, and synset coverage as objects to be placed in \holodeck scenes. Additional filtering was done for \procthorobja to keep only objects with synsets heuristically mapped to placement-compatible THOR categories, resulting in 92k objects across 2k synsets. Further details in filtering and curating objects from Objaverse are described in Appendix \ref{app:objects}.

Object models are accompanied by extensive physical and semantic metadata, convex colliders, and canonical coordinate definitions, besides grasps, which are obtained as described below. To support easy integration into robotics simulation workflows, all object models are provided in formats compatible with MuJoCo, IsaacSim, and ManiSkill.

\subsection{\graspdataset}

We introduce a comprehensive grasp dataset, \graspdataset, consisting of over 42 million grasps that cover objects in \name scenes. Our dataset provides 6-DoF grasp poses for two types of manipulable objects: rigid objects, which can be picked and moved as single bodies, and articulable objects, whose constituent parts can move relative to one another via a revolute or prismatic joint. Our grasp generation process builds on prior work such as GraspGen~\cite{Murali2025GraspGenAD}, 6-DOF GraspNet~\cite{mousavian2019graspnet} and ACRONYM~\cite{Eppner2020ACRONYMAL}, with extensions to support articulable objects through a new functionality-based evaluation step. We apply this pipeline to 48,111 objects drawn from a curated subset of Objaverse and custom-designed THOR assets, using separate pipelines for rigid and articulated objects to reflect their distinct manipulation requirements. Table~\ref{tab:grasp-comparison} compares our dataset with recent large-scale grasp datasets.

\begin{figure*}[t]
    \centering
    \includegraphics[width=0.99\linewidth]{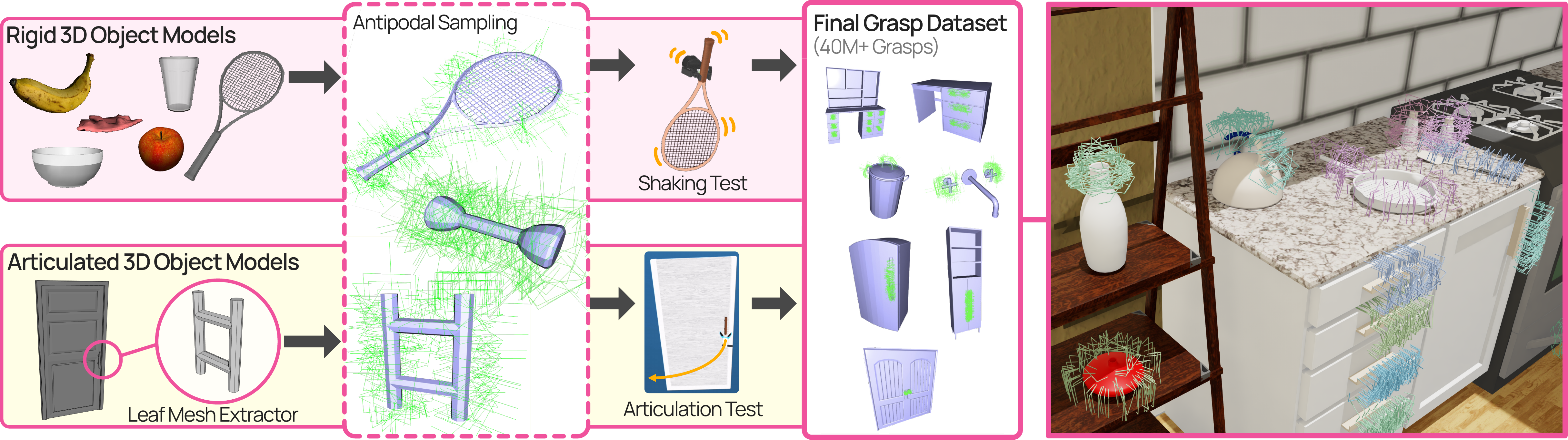}
    \caption{Our grasp generation pipeline consists of separate streams for rigid and articulated assets. We generate 42M+ verified grasps that can be utilized to create scripted interaction policies. Grasps can be used in different simulation environments, with an Issac example shown on the right.
    }
    \label{fig:grasp_pipeline}
\end{figure*}

\textbf{Grasp generation:} Our pipeline (Figure~\ref{fig:grasp_pipeline}) begins with 3D object models, from which we extract mesh colliders for grasp sampling. Antipodal contact pairs are sampled based on the geometry of the Robotiq 2F-85 gripper and the object. For rigid objects, sampling is performed across the full mesh surface, whereas for articulated objects, it is restricted to leaf components corresponding to handles or other functional interaction points. Grasps that result in collisions with non-leaf geometry are immediately discarded.

From the initial samples, we select up to 1,000 \textit{diverse} and \textit{robust} grasps per object. To ensure diversity, sampled grasps are clustered in the full 6-DoF pose space and tested uniformly across clusters. Grasp robustness is evaluated differently depending on the object type. For rigid objects, we apply controlled linear and rotational perturbations to the gripper upon grasping and discard any grasp that fails to maintain contact, leading to object slippage. For articulated objects, we define a grasp robustness by its actuation feasibility, meaning it can actuate the relevant joint through at least 70\% of its valid range in both directions, while maintaining stable contact.

To evaluate the practical utility of annotated grasps in MolmoSpaces scenes, we perform in-situ tests using a floating Robotiq 2F-85 gripper. Candidate grasps that collide with scene geometry are discarded, with collision checks performed at the pre-grasp pose (4 cm offset along the gripper’s negative local z-axis), grasp pose, and along the execution trajectory. The gripper then executes the pre-grasp and grasp motions, followed by lifting the object or articulating its moving part along the object’s joint path. Success rates (Fig.~\ref{fig:grasp_tests_thor_categories} in Appendix) are reported only for grasps that pass all collision checks and complete the intended motion.

\begin{figure}[htbp]
    \centering
    \includegraphics[width=0.4\columnwidth]{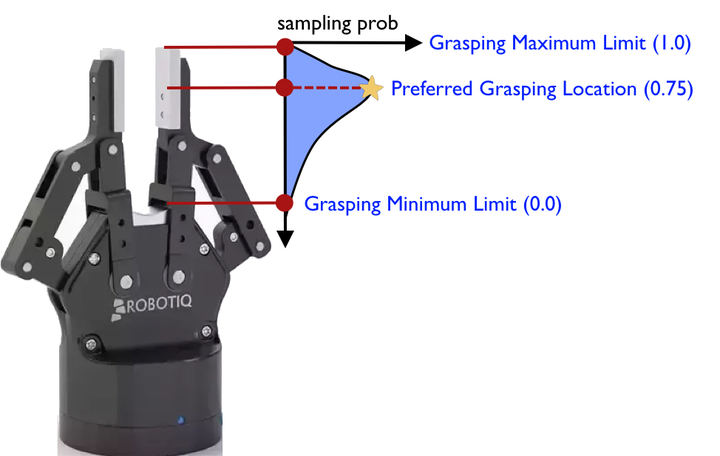}
    \caption[Short Title]{\centering RobotIQ 2F-85 gripper and preferred grasping locations for sampling grasps}
    \label{figure:gripper_contact}
\end{figure}

Initial evaluations showed that objects placed on surfaces often had few viable, non-colliding grasps. This is because grasp generation is performed on isolated objects without surrounding geometry and ignores gravity, resulting in high failure rates due to slippage during in-situ testing. To mitigate this, we updated the pipeline to generate more robust and stable grasps by biasing contacts toward the center of the fingertips as illustrated in Figure~\ref{figure:gripper_contact} . For small or thin objects, such as forks and pens, fingertip-edge contacts are preferred.

Remaining failures are largely due to scene- and object-level constraints: for lifting, common issues include objects too large for the lift pose, confined spaces limiting vertical clearance, collisions along the execution trajectory, and object slippage through the grasp; for articulated objects, failures occur when the object lifts instead of actuating, collisions exists along the articulation path, obstacles block articulation, or the gripper misaligns with the handle. These results underscore the importance of in-situ evaluation for producing grasps that are functional and practically useful given the way objects contextually exist within the scene.

\subsection{Robots} \label{sec:Robots}
We provide robots with varying amounts of mobility and complexity, covering the spectrum of widely-used manipulation platforms. We categorize these platforms as static or mobile, and single-arm or bimanual. To handle all of these cases, we provide a Franka FR3 arm with three gripper configurations (Franka Hand,  RobotIQ~2F-85, and CAP) as a static manipulator, Rainbow RB-Y1 as a bimanual and holonomic mobile manipulator, and floating \rum~\cite{cui2026contact} and RobotIQ grippers, unconnected to arms, so without kinematic limits. The Franka FR3 with the RobotIQ 2F-85 gripper is specifically set up as a DROID~\citep{khazatsky2024droid} system, with corresponding cameras with the correct intrinsics and extrinsics. The Franka FR3, RobotIQ Gripper, and Rainbow RB-Y1 robot models were sourced from~\cite{menagerie2022github}.

\textbf{Robot Control:} For manipulators, our framework provides for both absolute and relative joint position commands, which are tracked internally with a gravity-compensated joint-space stiffness controller. Mobile platforms, such as the holonomic RB-Y1 base or 6DoF floating \rum, can be controlled via absolute or relative poses.

\textbf{Kinematics Computation:} We provide built-in forward and inverse kinematics solvers for each robot, and the modular nature of our framework makes it easy to further extend for new robots. Our parallelized inverse-kinematics solver is written in JAX, and leverages Levenberg-Marquadt optimization with null-space control for posture regularization. This solver can natively be GPU-accelerated, but even on a CPU can solve batches as large as 256 samples with high precision in ${\sim}200$ms. For the RB-Y1 robot, which is configured for use with the cuRobo\cite{Sundaralingam2023CuRoboPC} motion generator, we provide forward and inverse kinematics as a wrapper around cuRobo's functionality.




\subsection{Modular experiment composition}

\begin{figure}[t]
    \centering
    \includegraphics[width=1.0\textwidth]{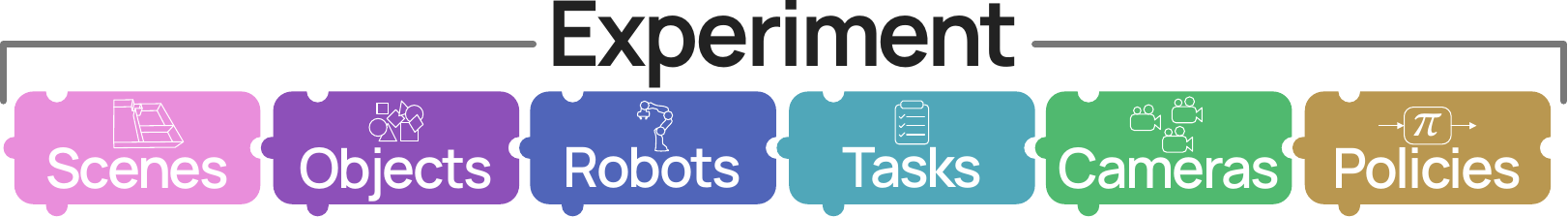} 
    \caption{Code Structure with modular experiment composition.}
    \label{fig:modular_experiment}
\end{figure}

\name supports modular experiment composition by flexibly combining scenes, tasks, robots, and camera configurations, as illustrated in the Figure~\ref{fig:modular_experiment}. We provide camera setups for commonly used RealSense, ZED, and GoPro cameras. Beyond vision inputs—including calibrated multi-view RGB and optional depth cameras with object image points—the framework exposes rich proprioceptive signals (e.g., joint states, end-effector and base poses), task-state information (object and articulation states), as well as task annotations, planner signals, and action histories. Together, these sensors capture the full interaction context.

\subsection{Data collection}

Two different modes of data collection are possible. Manual data collection is possible using TeleDex \cite{rayyan2026teledex} iPhone app, which uses iOS ARKit to stream the phone's pose. In addition to this, leveraging the pre-computed grasps, it would be possible for scripted policies to control the robot to solve the defined tasks.

\section{Benchmark}
To enable rigorous and reproducible evaluation of robot policies, we introduce \bench, which spans eight base tasks across navigation, manipulation, and mobile manipulation. These benchmarks are designed with explicit diversity requirements across scenes, object categories, and robot configurations, with each trial verified for solvability.
\subsection{Tasks}
\label{sec:benchmark_tasks}

\begin{figure*}[t]
\setlength{\tabcolsep}{2pt}      
\renewcommand{\arraystretch}{0.9} 
\footnotesize
\centering
\begin{tabular}{cccc}
\includegraphics[width=.24\linewidth]{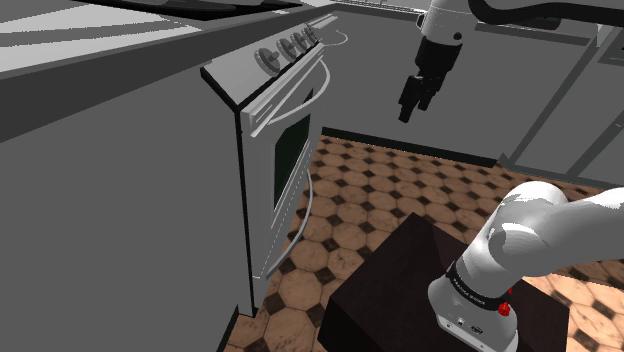} &
\includegraphics[width=.24\linewidth]{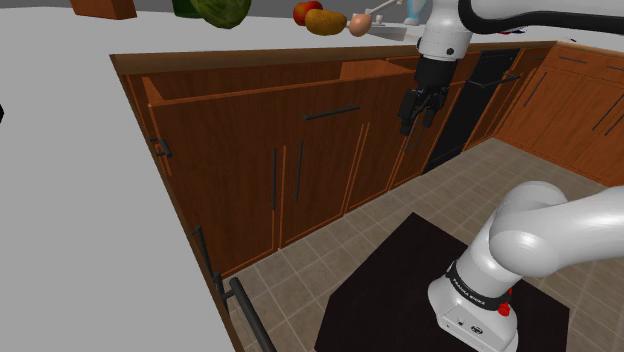} &
\includegraphics[width=.24\linewidth]{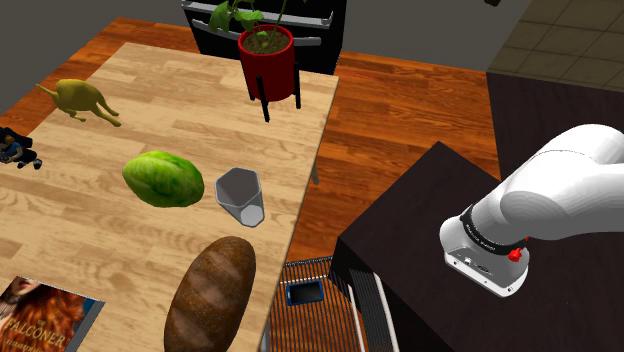} &
\includegraphics[width=.24\linewidth]{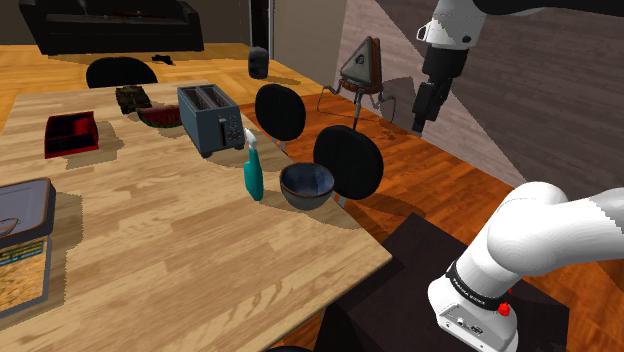} \\
a) ``open the oven" &
b) ``close the drawer" &
c) ``pick up the cup" &
d) ``place the spray in the bowl" \\
\includegraphics[width=.24\linewidth]{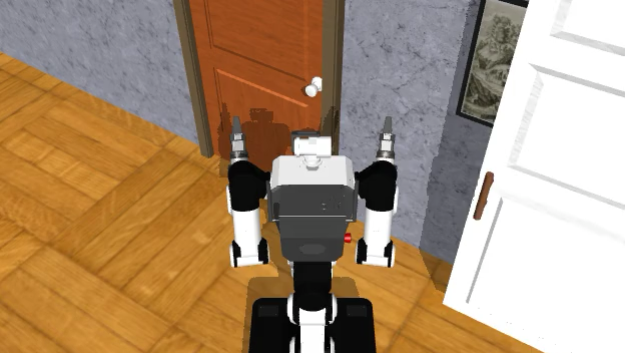} &
\includegraphics[width=.24\linewidth]{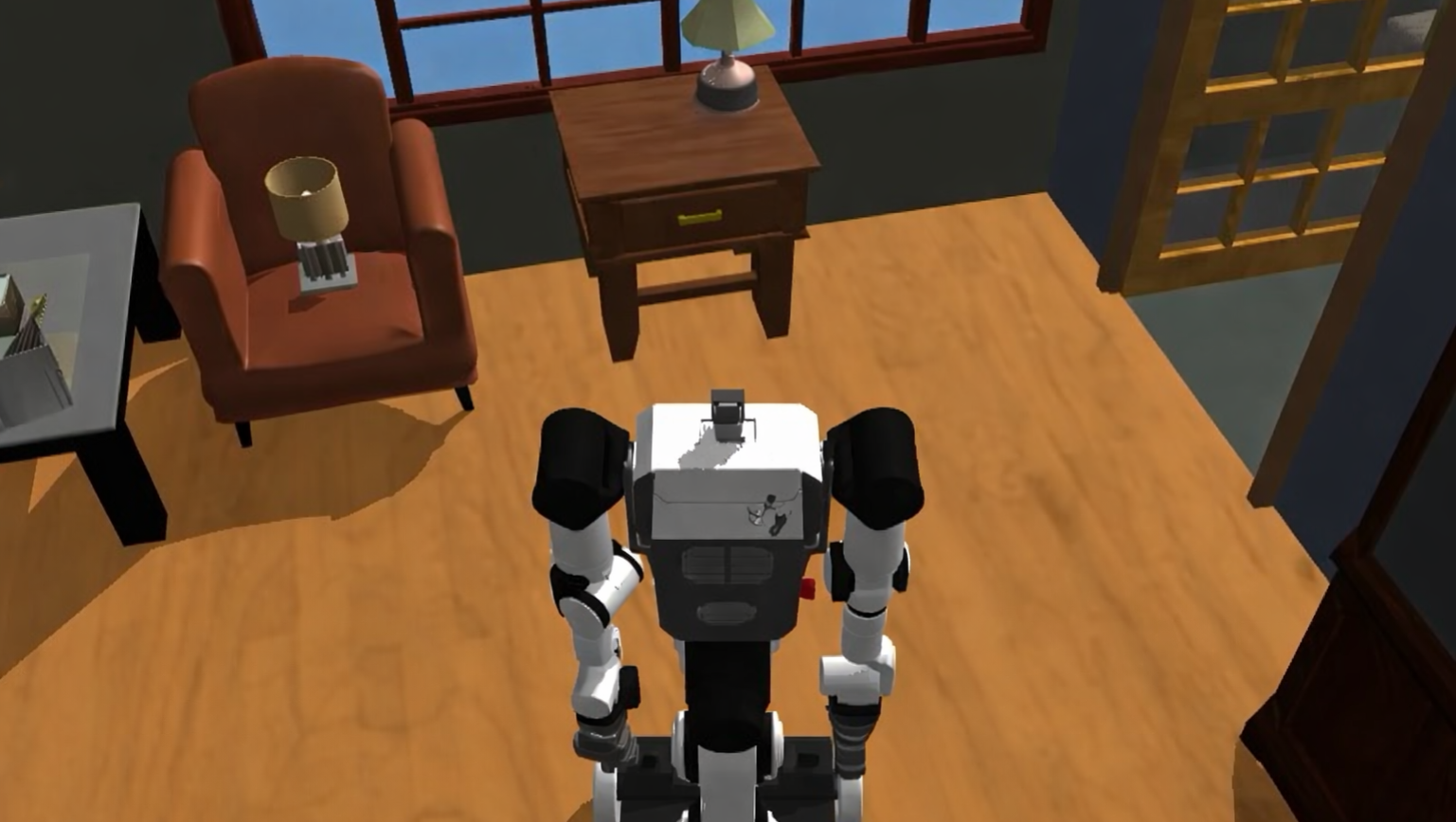} &
\includegraphics[width=.24\linewidth]{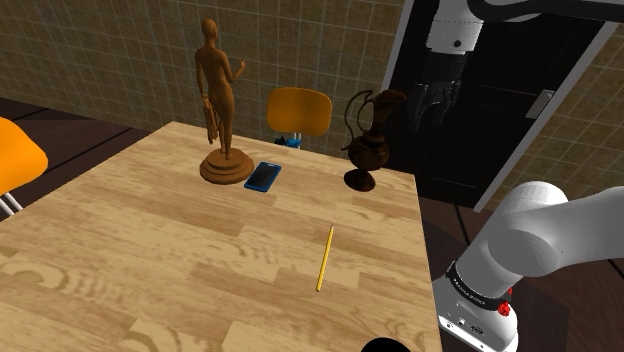} &
\includegraphics[width=.24\linewidth]{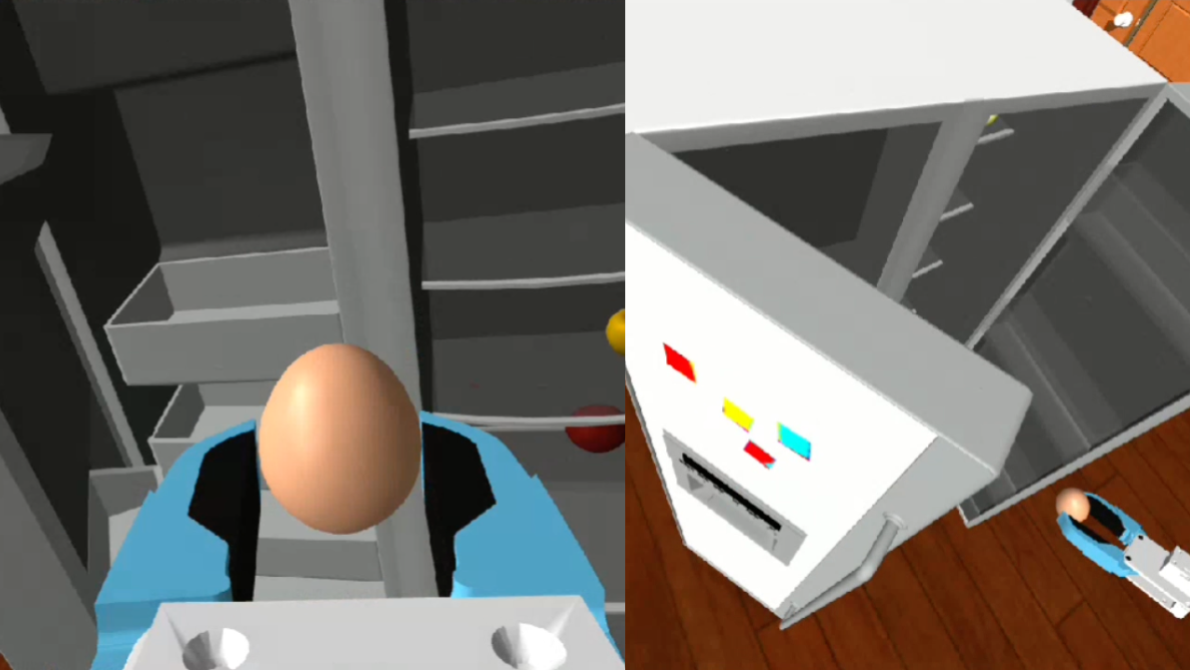} \\
e) ``pull" (door) &
f) ``find a lamp" &
g) ``place pencil next to vase" &
h) ``prepare a simple meal"\\
\end{tabular}
\caption{Example images of our range of tasks, spanning from manipulation of articulated and non-articulated assets to navigation and long-horizon tasks, shown together with their associated text instructions. These examples are from the MuJoCo simulator.}
\label{fig:task_examples}
\end{figure*}

Leveraging the diverse scenes, assets, and robots provided by \name, we introduce a suite of tasks and corresponding benchmarks designed for comprehensive policy evaluation. We define eight base tasks: \textit{navigate-to}, \textit{pick}, \textit{pick-and-place}, \textit{pick-and-place-next-to}, \textit{pick-and-place-color}, \textit{open}, \textit{close}, and \textit{open-door}. For each of these tasks, we also provide a well-defined success condition and a dense reward function.

\begin{enumerate}
    \item \textbf{Navigate-to:} A policy must search for and navigate to a specified target object. The robot is initialized between 4 and 20 meters away from the target object, possibly in an entirely different room. The success conditions require the object to be visible from the robot's head camera and closer than 1.5 meters away. We sample the same object candidate set as \cite{Ehsani2023SPOCIS}.
    
    \item \textbf{Pick:} Grasp and lift a specified object from its initial location by at least 1cm.
    \item \textbf{Pick-and-place:} Move a target object to be into or onto a target receptacle. To be counted as successful, at least 50\% of the object's weight must be vertically supported by the receptacle. Additionally, the target receptacle cannot have been displaced by more than 10 cm or 45 \degree{} from its initial pose.
    \item \textbf{Pick-and-place-color:} Similar to \textit{pick-and-place}, with the same initialization and success conditions. However, the task is complicated by multiple duplicates of the target receptacle differentiated only by color. This color is included in the task prompt, requiring policies to attend to and follow specific task instructions.
    \item \textbf{Pick-and-place-next-to:} Move a target object to be next to a target receptacle. The target receptacle is initialized 30-50 cm away from the target object, and to be successful, the policy must move the object to be closer than 5 cm surface-to-surface from the target receptacle. The object must additionally be on the same support surface (e.g., table) as the receptacle, which cannot have been displaced by more than 5 cm or 45\degree{} from its initial pose.
    \item \textbf{Open:} Open an articulated household fixture such as a drawer, cabinet, microwave, or fridge. Starting fully closed, the robot must open the fixture by at least 15\%.
    \item \textbf{Close:} Close an articulated household fixture, which is initialized halfway open. To succeed, the policy must close the fixture to at most 15\% open or 85\% closed. 
    \item \textbf{Open-door:} Open a hinged door by manipulating its handle or surface. The door starts fully closed, and the policy is tasked with opening it by at least 67\%.
\end{enumerate}

\paragraph{Navigation}
The \textit{navigate-to} task is evaluated with the Rainbow RB-Y1, which is instructed to locate and navigate to a given object. Following \cite{Ehsani2023SPOCIS}, the robot is initialized 4-20m away from the target object, and success is defined as being closer than 1.5m to the target object with it clearly visible in the navigation camera. For this task, policies must explicitly signal task completion, and incorrectly doing so is counted as a failure.

\paragraph{Rigid-body manipulation}
Non-articulated manipulation includes the \textit{pick}, \textit{pick-and-place}, \textit{pick-and-place-color}, and \textit{pick-and-place-next-to} tasks, which are evaluated with a Franka FR3 robot in the DROID configuration. For \textit{pick}, the robot must grasp and lift an object by at least 1cm. For \textit{pick-and-place}, the robot must move an object into or on a target receptacle, while \textit{pick-and-place-next-to} requires the robot to place the object next to the receptacle. The \textit{pick-and-place-color} task is a variant of \textit{pick-and-place} with multiple similar but differently-colored receptacles, requiring policies to attend and adhere to specific task instructions.

\paragraph{Articulated manipulation}
Our three articulated manipulation tasks cover both static and mobile manipulation.The \textit{open} and \textit{close} tasks are static manipulation tasks evaluated with the FR3 in the DROID setup, where the robot must open or close a variety of articulated household fixtures, including cabinets, drawers, refrigerators, and microwaves, by at least 15\%.The \textit{open-door} task is a mobile articulated manipulation task, where an RB-Y1 robot must push or pull a door to be at least 67\% open, requiring coordinated mobility and manipulation across many degrees of freedom.

Specifically, we evaluate following models: 

\begin{enumerate}
    \item \textbf{PI models (manipulation).}  
    We evaluate generalist models from the PI family, namely \texttt{$\pi_0$}, \texttt{$\pi_0$-FAST}, and \texttt{$\pi_{0.5}$}. As these models are trained primarily on real-world data, this setting constitutes a real-to-simulation evaluation. We follow the DROID hardware setup for this set of evaluations.

    \item \textbf{\rum models (manipulation).}  
    We evaluate \rum models that are task-specific and available for the following tasks: pick, open, and close. The models are conditioned on 3D ``contact'' points, which are provided by using Gemini-Robotics-ER-1.5~\cite{team2024gemini} at the initial step of the episode. We use the custom gripper design that \rum policies are trained with to replace the Robotiq Gripper from the DROID setup.
    
    \item \textbf{RING model (navigation).}  
    \texttt{RING} is an embodiment-agnostic indoor navigation policy trained entirely in simulation using large-scale randomization over robot body geometry and sensor configurations.
    
    \item \textbf{DualVLN (navigation).}
    DualVLN is a dual-system VLN foundation model that separates high-level reasoning from low-level control: a VLM-based global planner predicts mid-term waypoint goals, with a lightweight diffusion-based policy that executes smooth, real-time trajectories conditioned on these goals.

\end{enumerate}


\subsection{Benchmark creation}


For every task, we provide one or more benchmarks designed for comprehensive policy evaluation. These benchmarks draw from multiple scene datasets, described in Sec.~\ref{sec:scene_datasets}, and provide varying levels of difficulty and complexity. Concretely, each benchmark is defined by an initial scene, robot, and camera configuration, as well as a descriptive task instruction. To ensure benchmark quality, we perform balancing to maximize diversity of object categories and instances, as well as scenes. Additionally, each of our provided benchmarks are guaranteed to be solvable with the provided robot and initial conditions, ensuring task feasibility.

We generate a full set of benchmarks with all combinations of environments and tasks, listed in Table~\ref{tab:all_benchmarks}. For easier comparisons, we also include preferred evaluation configurations. For the open and close tasks, we choose the MSCrafted scenes, as these contain handcrafted kitchens with many cabinets and drawers. For most manipulation tasks, we choose the MSProcObj variants, as the presence of objaverse assets gives object diversity. For the door-opening task, we use the MSProc environment, as object diversity is irrelevant. All tasks have 2k samples. We present select benchmark results in Sec.~\ref{sec:evaluations}.

\newcommand{\cmark}{$\color{ForestGreen}\textbf{\ding{51}}$}

\begin{table*}[t]
\centering
\small
\setlength{\tabcolsep}{4pt}
\begin{tabular}{lcccc}
\toprule
\textbf{Task} & \textbf{MSCrafted} & \textbf{MSProc} & \textbf{MSProcObj} & \textbf{MSMultiType} \\
\midrule
Open (Franka) & \cmark+e &  $\checkmark$ & $\checkmark$ & $\checkmark$ \\
Close (Franka) & \cmark+e &  $\checkmark$& $\checkmark$ & $\checkmark$ \\
Pick (Franka) & $\checkmark$ & $\checkmark$+e & \cmark & $\checkmark$ \\
Pick \& Place (Franka) & $\checkmark$ & $\checkmark$+e  & \cmark  & $\checkmark$ \\
Pick \& Place Next To (Franka) & $\checkmark$ & $\checkmark$ & \cmark & $\checkmark$ \\ 
Door Open (RBY1) & $\checkmark$ & \cmark & $\checkmark$ & $\checkmark$ \\
Pick (RBY1) & $\checkmark$ & $\checkmark$ & \cmark & $\checkmark$ \\
Pick \& Place (RBY1) & $\checkmark$ & $\checkmark$ & \cmark & $\checkmark$ \\
Navigation (RBY1) &  &  & $\checkmark$ & \cmark \\

\bottomrule
\end{tabular}
\caption{Availability of manipulation and navigation benchmarks across environments. $\checkmark$ indicates the benchmark is available, +e indicates there is an easy variant of the benchmark, \cmark indicates the preferred evaluation environemnt.}
\label{tab:all_benchmarks}
\end{table*}

\subsection{Extensions}
Controlled variants of benchmarks enable easier testing of hypotheses. For example, to answer the question ``Does a policy perform as well on open vocabulary navigation as on closed vocabulary navigation?", one can evaluate on \procthor, which uses a closed set of object categories, versus \holodeck, which supports open vocabulary. In addition to this, \name also allows for the easy creation of controlled adversarial benchmarks to test specific functionality, such as the ability of a policy to handle varying initial robot configurations (shown in \cref{fig:robustness_perturbations}) or the ability to do manipulation in progressively more cluttered environments. 

\begin{figure}[h]
    \centering
    \includegraphics[width=1\linewidth]{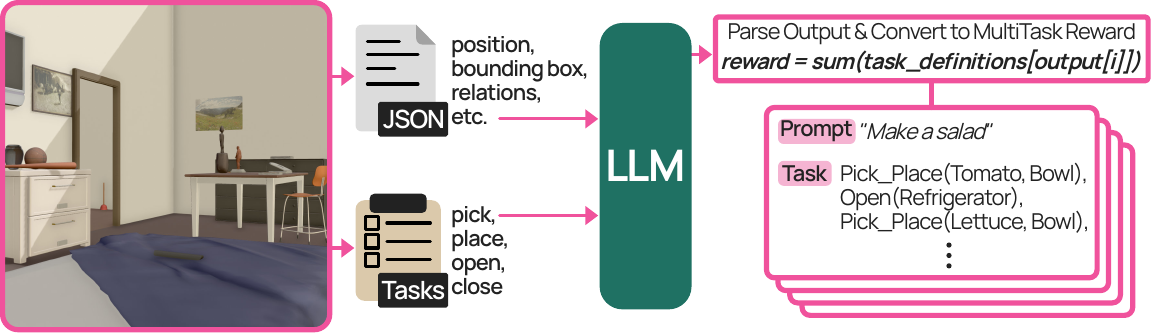}
    \caption{The LLM-based long-horizon task generation system makes use of text-form scene descriptions to generate new task descriptions and success condition checks based on predefined atomic checks.}
    \label{fig:llm_task}
\end{figure}

Additionally, while our provided benchmarks cover a variety of short-horizon tasks, long-horizon evaluation is also a critically important aspect of benchmarking. To that end, we provide an LLM-based task generation system that can generate \textbf{long-horizon tasks} with more abstract task descriptions. In this process, shown in \cref{fig:llm_task}, we take an existing scene and prompt an LLM to generate feasible long-horizon tasks, defined by a high-level task description and a sequence of base tasks as defined in Sec.~\ref{sec:benchmark_tasks}. Unlike the base tasks, the tasks obtained by this sampling procedure can result in longer horizons, e.g., needing to open a fridge before taking objects out and placing them elsewhere.

\begin{figure*}[t]
    \centering
    \includegraphics[width=0.99     \linewidth]{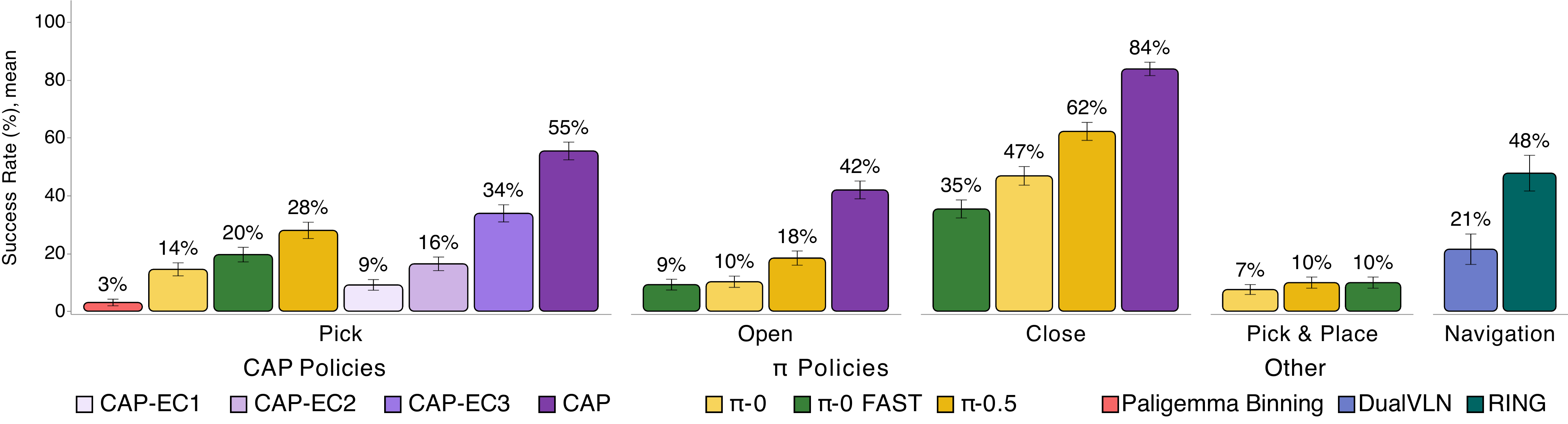}
    \caption{Zero-shot success rates of different baseline policies across five \name benchmark tasks. Showing expected performance improvement in improved policies. Error bars show 95\% Bayesian credible intervals. }
    \label{fig:model_comparison}
\end{figure*}

\section{Experiments}

\subsection{Evaluations}
\label{sec:evaluations}
We use \bench to evaluate open-source manipulation and navigation policies across the diverse scenes of \name zero-shot, without task-specific fine-tuning as done in many prior evaluations. We believe this setting is more practical, as it evaluates models as they are released and intended to be used, rather than their ability to efficiently fit additional task-specific data. We report the results under the standard settings in in \cref{fig:model_comparison}.

\subsubsection{Manipulation tasks} We evaluate vision–language–action (VLA) models from the $\pi$ family \cite{black2026pi0visionlanguageactionflowmodel,pertsch2025fast,intelligence2025pi05visionlanguageactionmodelopenworld} ($\pi_0$, $\pi_0\text{-FAST}$, and $\pi_{0.5}$, specifically the joint position variants that have been fine-tuned on the DROID dataset \cite{Jain2025PolaRiSSR}) as well as utility contact-action models from \rum \cite{cui2026contact}. These models make use of different setups as described in \cref{sec:Robots}, and therefore operate under different sensing and embodiment constraints. In particular, the DROID setup provides the $\pi$ models with two camera views, whereas the floating CAP setup uses a single wrist-mounted camera and has no kinematic constraints. During benchmark construction, we ensure that all manipulation tasks are feasible under the DROID setup. 

\subsubsection{Navigation Tasks} We select two state-of-the-art prior methods for visual object-goal navigation. RING~\cite{Eftekhar2024TheOR} is an embodiment-agnostic, transformer-based navigation policy trained entirely in simulation that demonstrates robust generalization across a wide range of real-world robot platforms. It is trained on semantic navigation tasks where instructions consist of a simple verb (e.g., ``go to", ``locate", ``find", ``search for", ``navigate to") paired with an object category. DualVLN~\cite{Wei2025GroundSM} is a dual-system vision–language navigation (VLN) foundation model that integrates high-level reasoning with low-level action execution. Unlike RING, DualVLN is trained on detailed instructions that specify intermediate steps the robot must follow, rather than simple semantic goals (see Fig. 4 of \cite{Wei2025GroundSM} for examples). Both policies are evaluated on the \textit{navigate-to} benchmark, which comprises 2,000 trajectories across 679 houses and goal objects drawn from 568 synset categories. This difference in task formulation explains the performance gap between policies (see Fig.~\ref{fig:model_comparison}), as our benchmark's semantic navigation format poses a distributional mismatch for DualVLN while aligning with RING's training. We adopt semantic navigation because it is more directly compatible with downstream mobile manipulation tasks.

\begin{figure*}
    \centering
    \includegraphics[width=\linewidth]{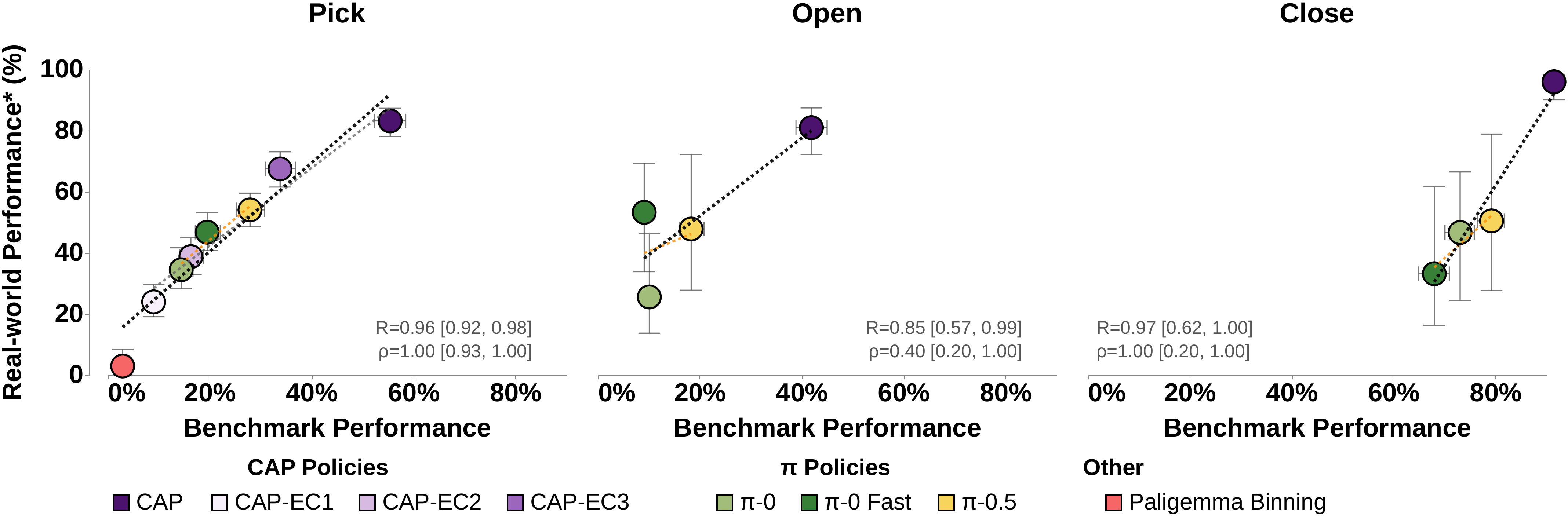}
    \caption{Sim-to-real correlation results for \textit{pick}, \textit{open}, and \textit{close}  task. Coefficient of determination ($R$) and the Spearman rank correlation coefficient ($\rho$) are shown.}
    \label{fig:sim2real_all}
\end{figure*}



\subsection{Sim-to-Real Correlation}

Correlation between performance in real and simulated evaluations shows how predictive simulation results are of real world behavior.
We therefore compare the results achieved by policies in our benchmarks to their real-world performance, which we take from RoboArena \cite{Atreya2025RoboArenaDR} and \rum \cite{cui2026contact}. We evaluate the correlation for the \textit{pick}, \textit{open}, and \textit{close} tasks individually. Results for are shown in \cref{fig:sim2real_all}. For the \textit{pick} task, we observe a strong linear correlation between our \bench{} results and the results from 752 RoboArena \textit{pick} tasks, with Pearson and Spearman rank correlation coefficients of 0.96 and 0.98, respectively. This underlines \bench{}'s utility and predictive power. As in the Pick task, we compute correlations between benchmark success rates and real-world performance measured on RoboArena and CAP for the \textit{open} and \textit{close} tasks. While fewer real-world episodes are available for these tasks, we observe consistent positive correlations but with bigger error bars. 
    
\begingroup
\renewcommand{\thefootnote}{\fnsymbol{footnote}}%
\footnotetext[1]{The real-world performance is obtained from RoboArena by filtering for pick tasks and using RoboArena’s partial success criteria.}
\endgroup

\subsection{Distributional Evaluations}

Machine learning models typically perform best under small distribution shifts between training and evaluation. Given that the $\pi$ series models are trained on datasets that include DROID~\cite{khazatsky2024droid}, we evaluate how performance degrades as we move away from this distribution. This evaluation is enabled by the scale of \name, which allows us to systematically vary environmental and policy factors beyond aggregate success metrics.





\begin{figure*}[t]
    \centering
    \begin{minipage}{0.48\textwidth}
        \centering
        \includegraphics[width=\linewidth]{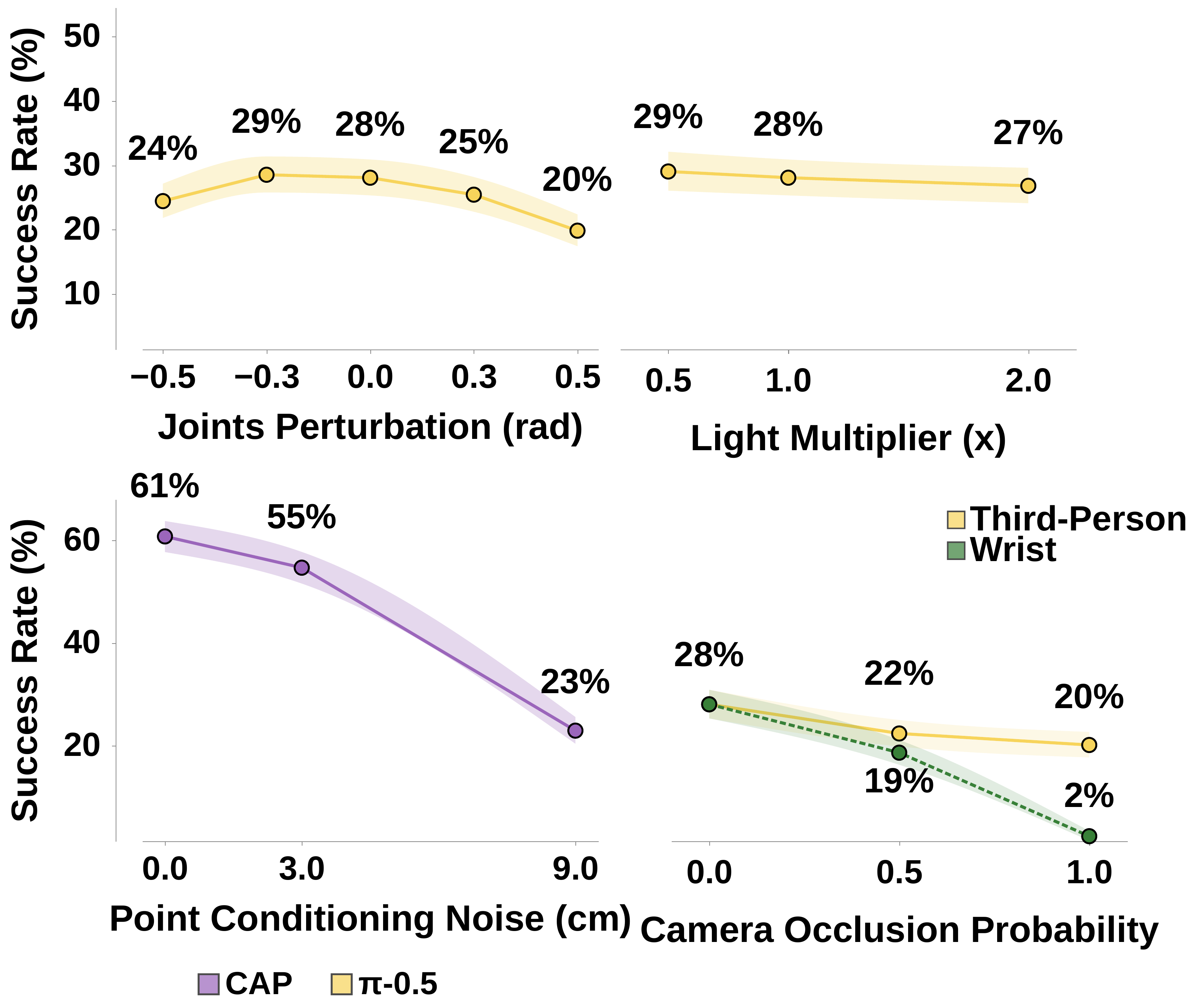}
        \caption{Robustness analysis under environmental perturbations. \textbf{Top:} Joint noise and lighting. \textbf{Bottom:} Point noise and camera occlusion.}
        \label{fig:robustness_perturbations}
    \end{minipage}
    \hfill
    \begin{minipage}{0.48\textwidth}
        \centering
        \includegraphics[width=\linewidth]{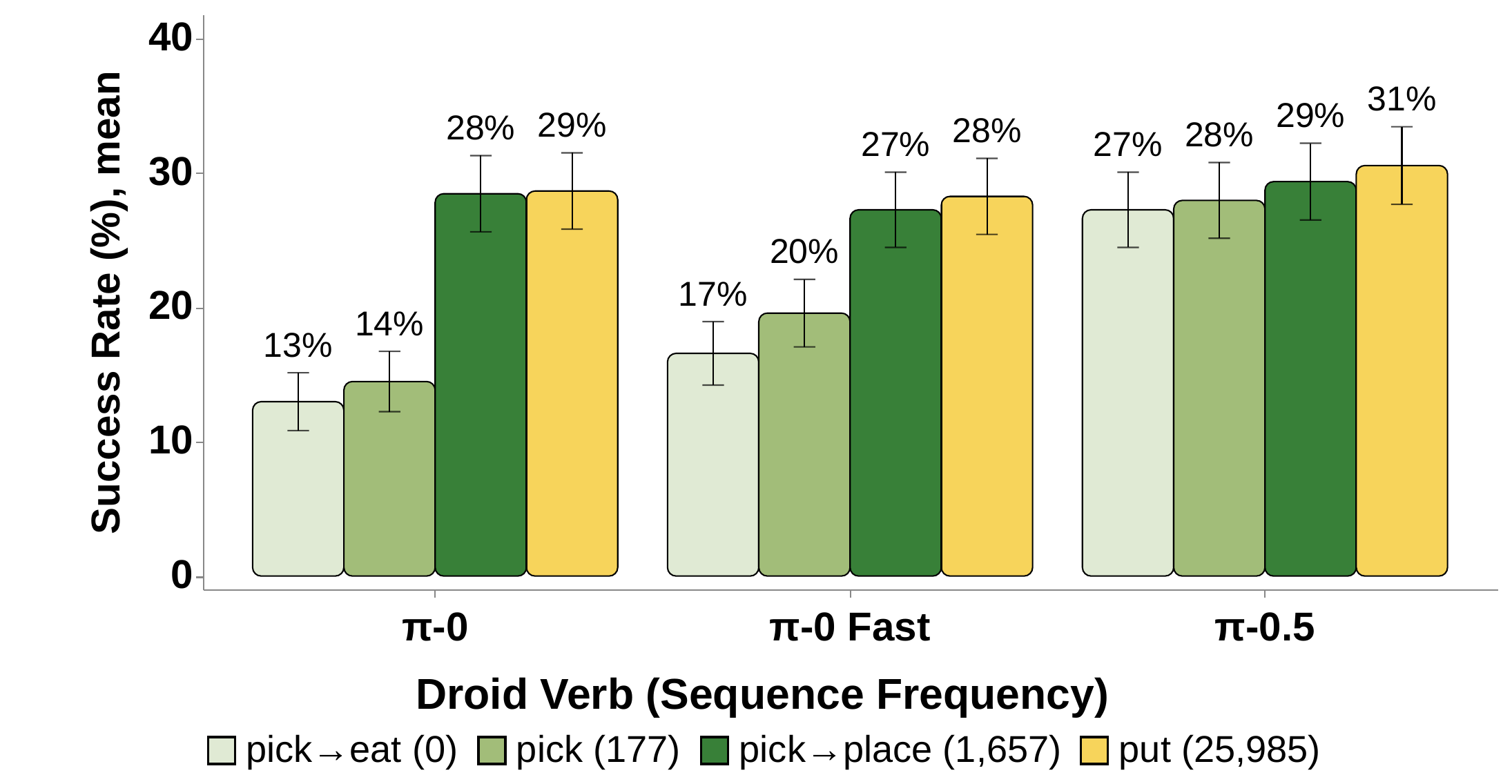}
        \caption{Prompt sensitivity: $\pi$ models performance on the Pick-MSProc-1k benchmark with varying prompts. Frequent DROID prompts perform better.}
        \label{fig:eval_protocol_d}
    \end{minipage}
\end{figure*}

To probe language-induced distribution shift, we vary the \textit{pick} task prompt using verb sequences of increasing frequency in DROID instructions, as shown in Figure~\ref{fig:eval_protocol_d}. When using phrasing that are more frequent in the DROID dataset, $\pi_0$ achieves a success rate within 1\% of $\pi_{0.5}$, compared to a 14\% gap otherwise. With the assumption that the underlying DROID data distributions used to fine-tune $\pi$ models are similar, this sensitivity to prompt phrasing suggests that generalization limitations of earlier $\pi$ models arise from the language-conditioning component rather than the action heads. We observe this effect for the \textit{pick} task in our benchmark and leave validation of its generality in other tasks to future work.

A similar degradation is observed when perturbing the initial joint positions of the $\pi_{0.5}$ policy from the default configuration used in DROID. As shown in Figure~\ref{fig:robustness_perturbations}, varying the starting joint positions away from this default leads to a decline in performance. In contrast, varying the lighting intensity has little effect on performance, likely because such visual distribution shifts are mitigated through image-based data augmentation.

\begin{figure*}
    \centering
    \includegraphics[width=0.9\linewidth]{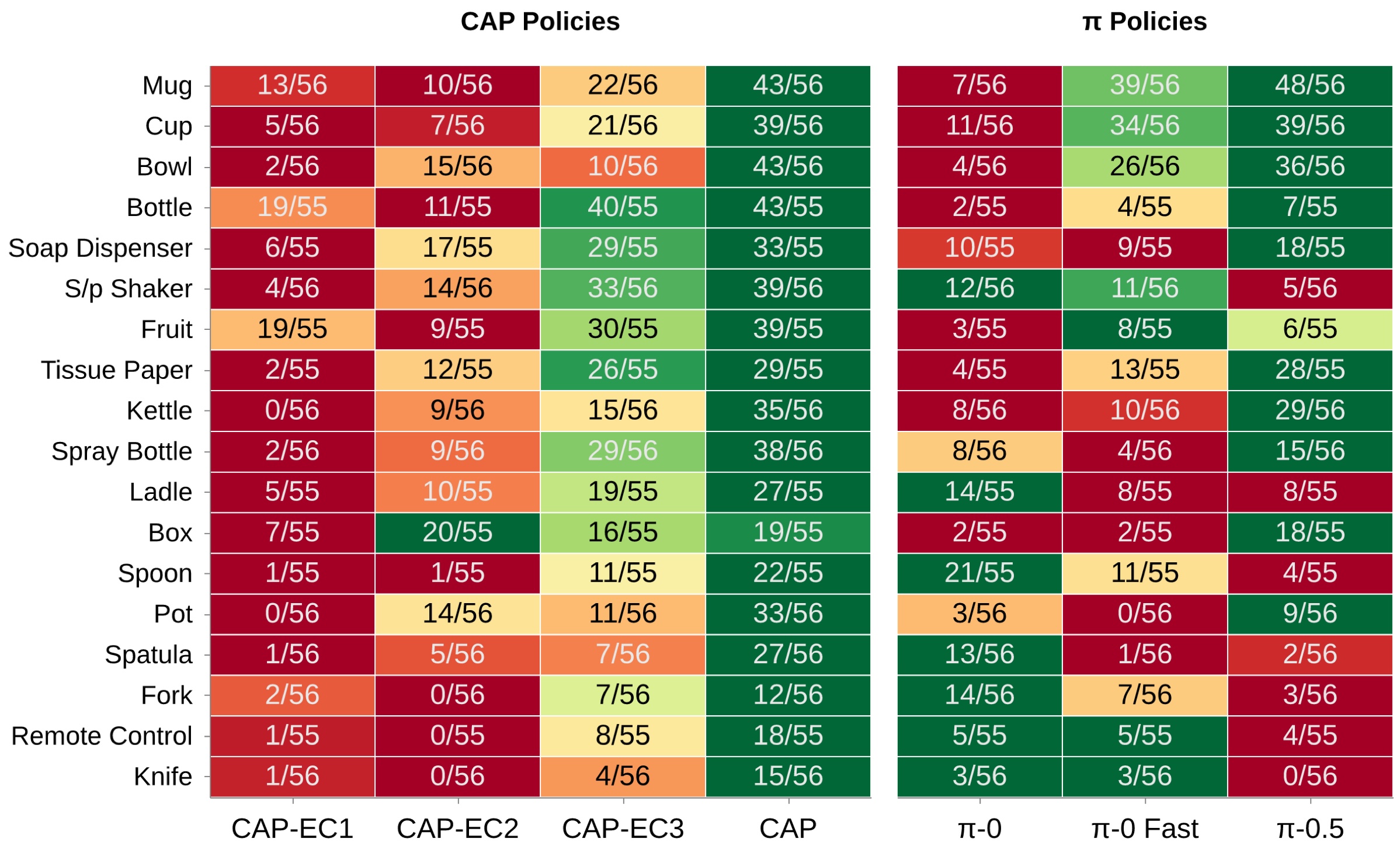}
\caption{Per-object category success rates for CAP and $\pi$ policy families on the pick task. Colors are normalized per row within each table.}

    \label{fig:object_sr}
\end{figure*}

Figure~\ref{fig:robustness_perturbations} also illustrates the effect of other environmental perturbations on the performance of policies. In particular, occluding the wrist camera reduces $\pi_{0.5}$ success rate to 2\%, while occluding the third-person camera only lowers the performance to 20\%, indicating a strong reliance on wrist-mounted visual input. Similarly, CAP's performance indicates a strong reliance on a good starting conditioning point.
Figure~\ref{fig:grasp_histogram} shows that $\pi_{0.5}$ and \rum{} prefer different grasping approaches, with $\pi_{0.5}$ favoring top-down grasps and \rum{} favoring side grasps. This preference helps explain why $\pi_{0.5}$ performs better on objects with top openings, such as mugs and bowls, while \rum{} performs better than $\pi_{0.5}$ on objects where side grasps are feasible, such as bottles and apples, as shown in Figure~\ref{fig:object_sr}.


\subsection{Controlled Policy Comparison}
\label{sec:policy_comparison}
To ensure a fair comparison between policies, we use a Franka FR3 arm setup as a unified embodiment for all manipulation benchmarks, alongside three RGB-D cameras: one wrist-mounted and two third-person. We allow a fixed offset in the robot base frame relative to the initial task configuration to accommodate grippers with differing geometries. This offset must be constant across episodes and should not use any privileged or task-specific information. During task generation, we filter out benchmark tasks that are not physically achievable under the standard DROID setup. We acknowledge that this filtering procedure may advantage grippers such as the RobotIQ 2F-85. We also acknowledge that the choice of end-effector can affect task difficulty, but consider this a part of system design to be evaluated.

\begin{figure}[t]
    \centering
    \begin{minipage}[b]{0.48\linewidth}
        \centering
        \includegraphics[width=\linewidth]{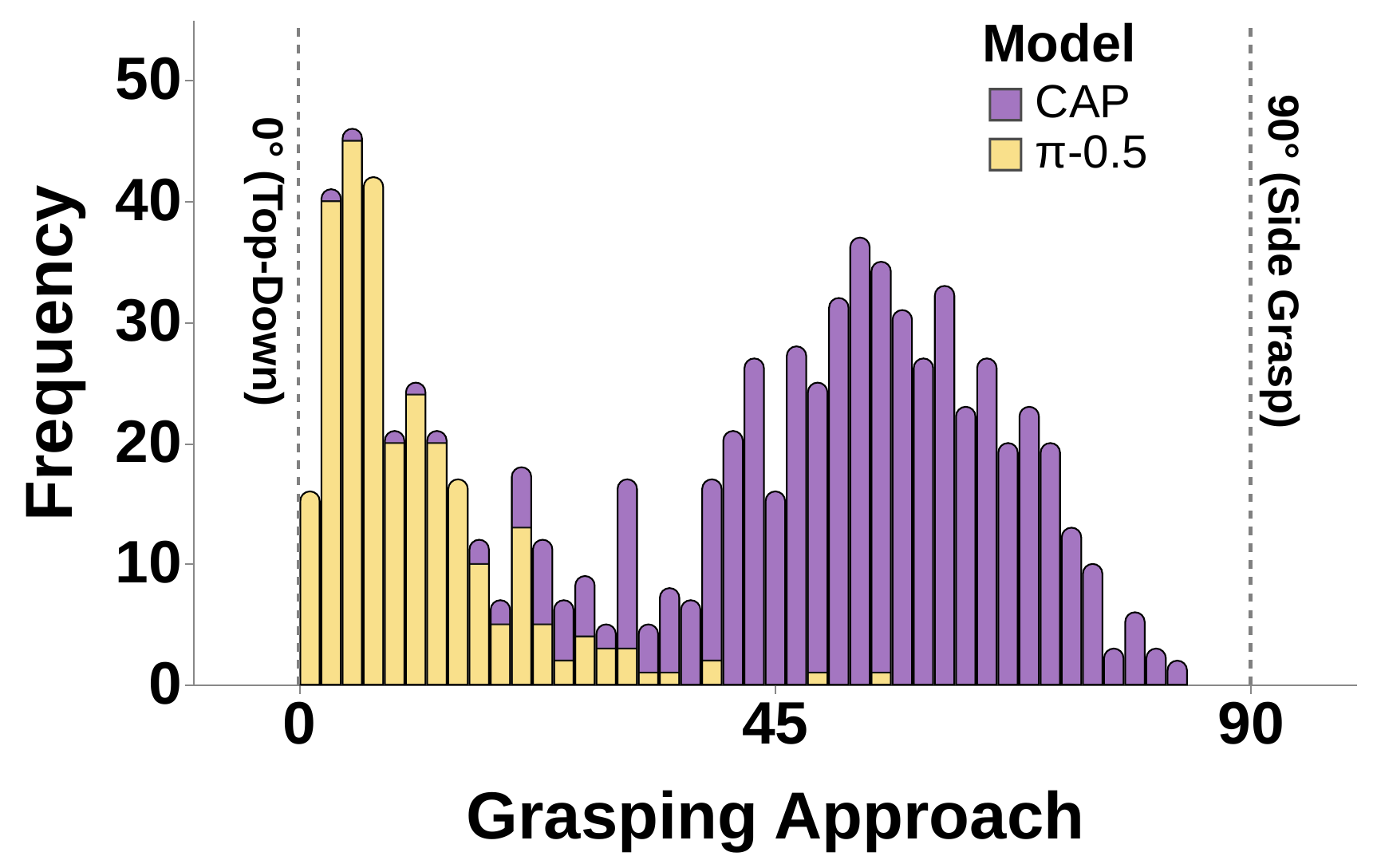}
        \caption{Grasp direction histogram of successful grasp of the $\pi_{0.5}$ and \rum{} policies.}
        \label{fig:grasp_histogram}
    \end{minipage}
    \hfill
    \begin{minipage}[b]{0.48\linewidth}
        \centering
        \includegraphics[width=\linewidth]{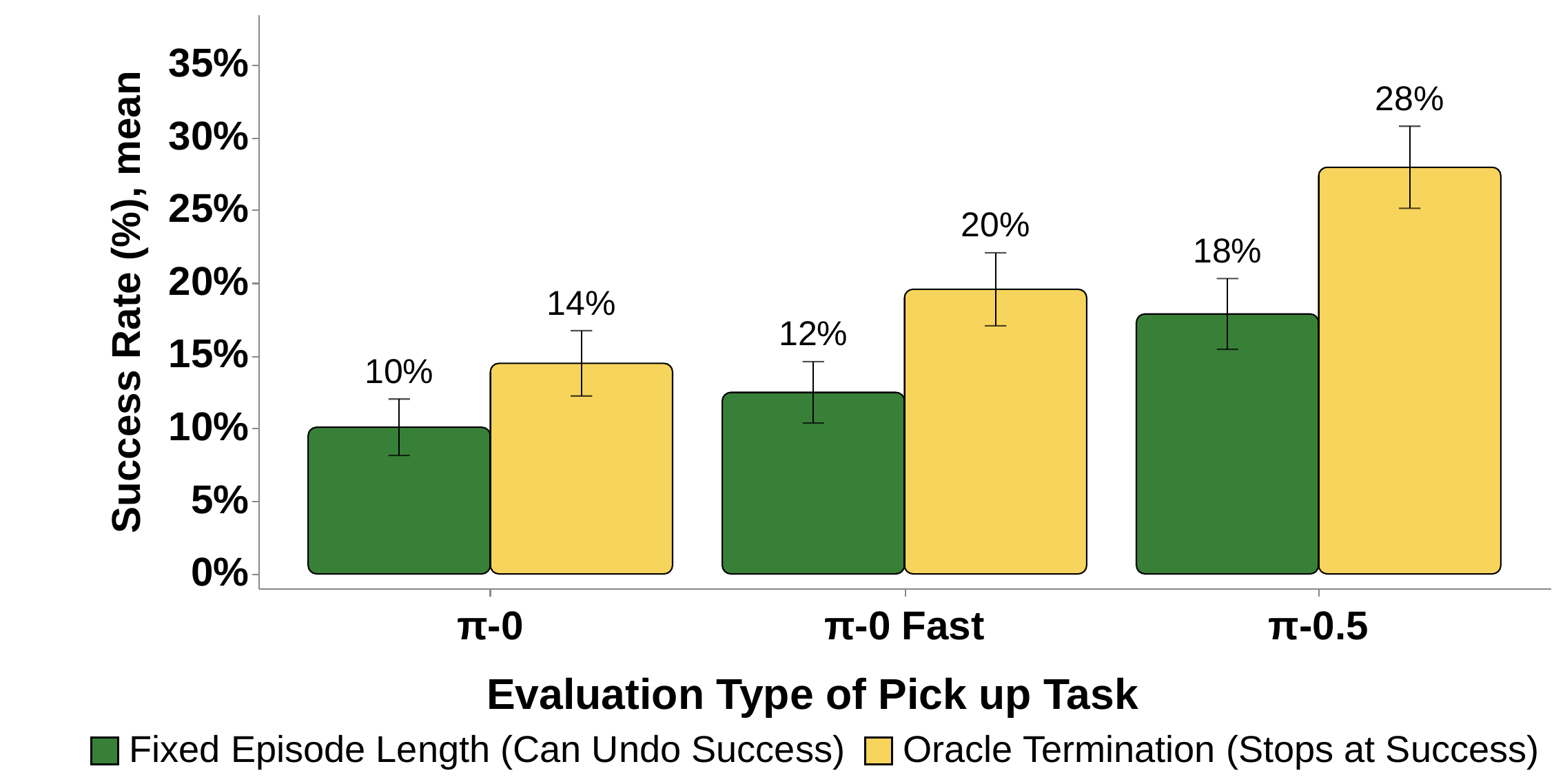}
        \caption{Comparison of oracle termination and fixed-horizon termination.}
        \label{fig:oracle}
    \end{minipage}
\end{figure}

Another variable that affects our recorded policies performance is the task horizon, which we set as 300 for the $\pi$ models, and 50 for CAP. Figure~\ref{fig:oracle} compares oracle termination with fixed-horizon termination of the $\pi$ models, showing that policies sometimes reach a successful state but subsequently undo it before the episode ends. This behavior is also reflected in Table~\ref{tab:grasp_transitions}, where on average $\pi_{0.5}$ goes through 2.65 grasp-ungrasp transitions before exceeding the reward threshold in successful episodes, while $\pi_0$ makes 4.63 such transitions. These results suggest that $\pi$ models benefit from sufficient time to retry actions, whereas CAP relies on VLM-based supervision for retries that is not used in our baseline benchmark experiments.

\begin{table}[h]
\centering
\caption{Average number of grasp-ungrasp transitions before task success for $\pi$ policies on the pick task.}
\label{tab:grasp_transitions}
\begin{tabular}{lccc}
\toprule
Policy & Mean $\pm$ Std & Median \\
\midrule
$\pi$-0              & $4.63 \pm 5.26$ & 3.0 \\
$\pi$-0 Fast         & $2.02 \pm 2.62$ & 1.0 \\
$\pi$-0.5            & $2.65 \pm 2.74$ & 1.0 \\
\bottomrule
\end{tabular}
\end{table}

\section{Conclusion} 
\label{sec:conclusion}

We present \name, a comprehensive open ecosystem comprising large-scale simulation environments, diverse object assets, and extensive grasp datasets. We further introduce \bench, a benchmark suite spanning base skills and LLM-assisted long-horizon tasks, with controlled difficulty and perturbations that enable detailed analysis and principled comparison of generalist robot policies. We validate \bench's utility with thorough experimentation, demonstrating strong sim-to-real correlation and benchmarking multiple SOTA policies, including distributional evaluations that reveal subtle policy behavior characteristics.
Although simulation remains inherently imperfect, well-designed simulation benchmarks provide a critical foundation for evaluating robotic policies and guiding progress toward robust real-world performance. In the future, we plan to support data generation and reinforcement learning for robot policies at scale, enabling us to further study scaling behaviors for robot foundation models.

\clearpage

\section*{Author Contributions}
\label{sec:contrib}

This project was made possible through the equal contributions of all four co-first authors, in no particular order. Their individual contributions are as follows:
\begin{itemize}
    \item \textbf{Yejin Kim}: Led the project and initially converted all assets and scenes from AI2-THOR to MuJoCo MJCF format; contributed to asset quality testings, grasp generation pipeline and evaluations, benchmark creation, and baseline evaluations.
    \item \textbf{Wilbert Pumacay}: Led the asset conversion and quality testings across MuJoCo, Isaac and ManiSkill simulators.
    \item \textbf{Omar Rayyan}: Led distributional benchmark evaluations, the addition of $\pi$, CAPs baselines and teleoperation data-collection; led the grasp generation pipeline and evaluations; contributed to benchmark creation.
    \item \textbf{Max Argus}: Led the benchmark creation and baseline evaluations.
\end{itemize}

All other contributors are also deeply appreciated for their effort, which is critical to the success of the \name project. As not all of these can be captured, we indicate their primary contributing role in \name:
\begin{itemize}
    \item For assets conversion and physical parameter tunings: Yejin Kim, Wilbert Pumacay, Winson Han, Eli VanderBilt, Jordi Salvador,  and Shuo Liu.
    \item For grasp generation and evalautions: Omar Rayyan, and Yejin Kim.
    \item For benchmark creation and baseline evaluation: Max Argus, Omar Rayyan, Yejin Kim, Arjun Guru, Maya Guru, Abhay Deshpande, Rose Hendrix, Snehal Jauhri, Shuo Liu, Ainaz Eftekhar, Ying-Chun Lee, and Piper Wolters. Nur Muhammad Mahi Shafiullah advised the zero-shot and distributional evaluations.
    \item For multi-simulator support: Wilbert Pumacay, Alvaro Herrasti, and Donovan Clay
    \item For paper writing and figures:  Yejin Kim, Max Argus, Wilbert Pumacay, Omar Rayyan, Winson Han, Eli VanderBilt, Abhay Deshpande, Rose Hendrix, Maya Guru, Arjun Guru, Jordi Salvador, Jiafei Duan, Nur Muhammad Mahi Shafiullah, and Ranjay Krishna.
    \item For project management: Karen Farley.
    \item For research advisory: Ranjay Krishna, Dieter Fox, and Ali Farhadi.
    \item Project PI: Ranjay Krishna 
\end{itemize}

\section*{Acknowledgment}

This work would not be possible without the support of our colleagues at Ai2:

\begin{itemize}
    \item We thank Rachel Ratner and Karen Goodfellow for the creation and the support of robot demo webpage for \name.
    \item We thank David Albright, Crystal Nam, Kristin Cha, Sophie Lebrecht, Taira Anderson, Kyle Wiggers, Kelsey MacMillan, Katie Morigi, and Megan Bartot for project management, support to robot room and publicity of \name
    \item We thank Yoganand Chandrasekhar, Johann Dahm, Fangzhou Hu, and Caroline Wu for their work on the Ai2 cluster.

\end{itemize}

\clearpage

\bibliographystyle{unsrtnat}  
\bibliography{references}
\clearpage

\appendix
\appendix
\section*{Appendix}

\section{Object Model Dataset Details}
\label{app:objects}

\begin{figure*}[htb]
    \centering
    {\footnotesize\fontfamily{cmss}\selectfont
    \begin{tabular}{cc}
    \includegraphics[width=0.43\linewidth]{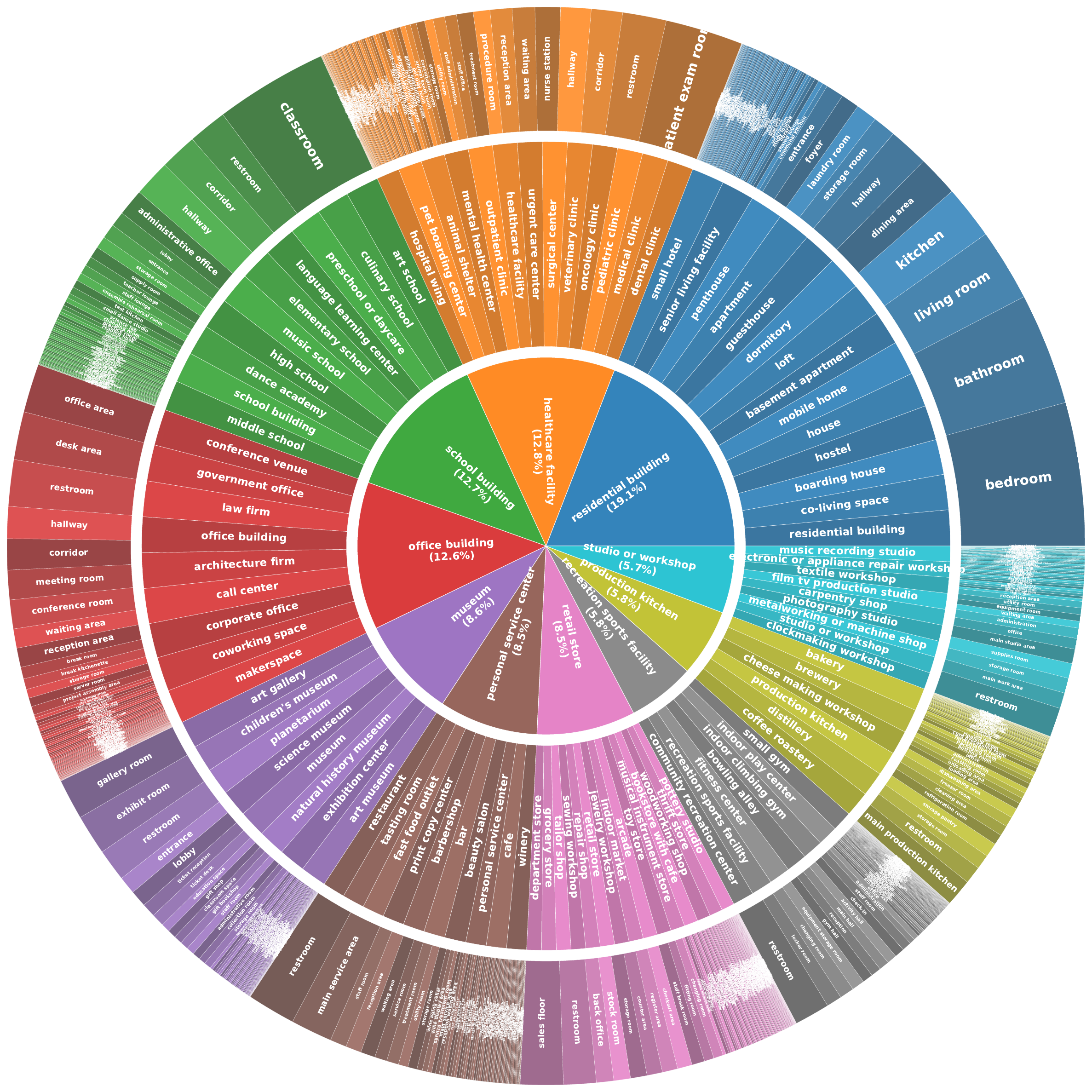} &
    \includegraphics[width=0.43\linewidth]{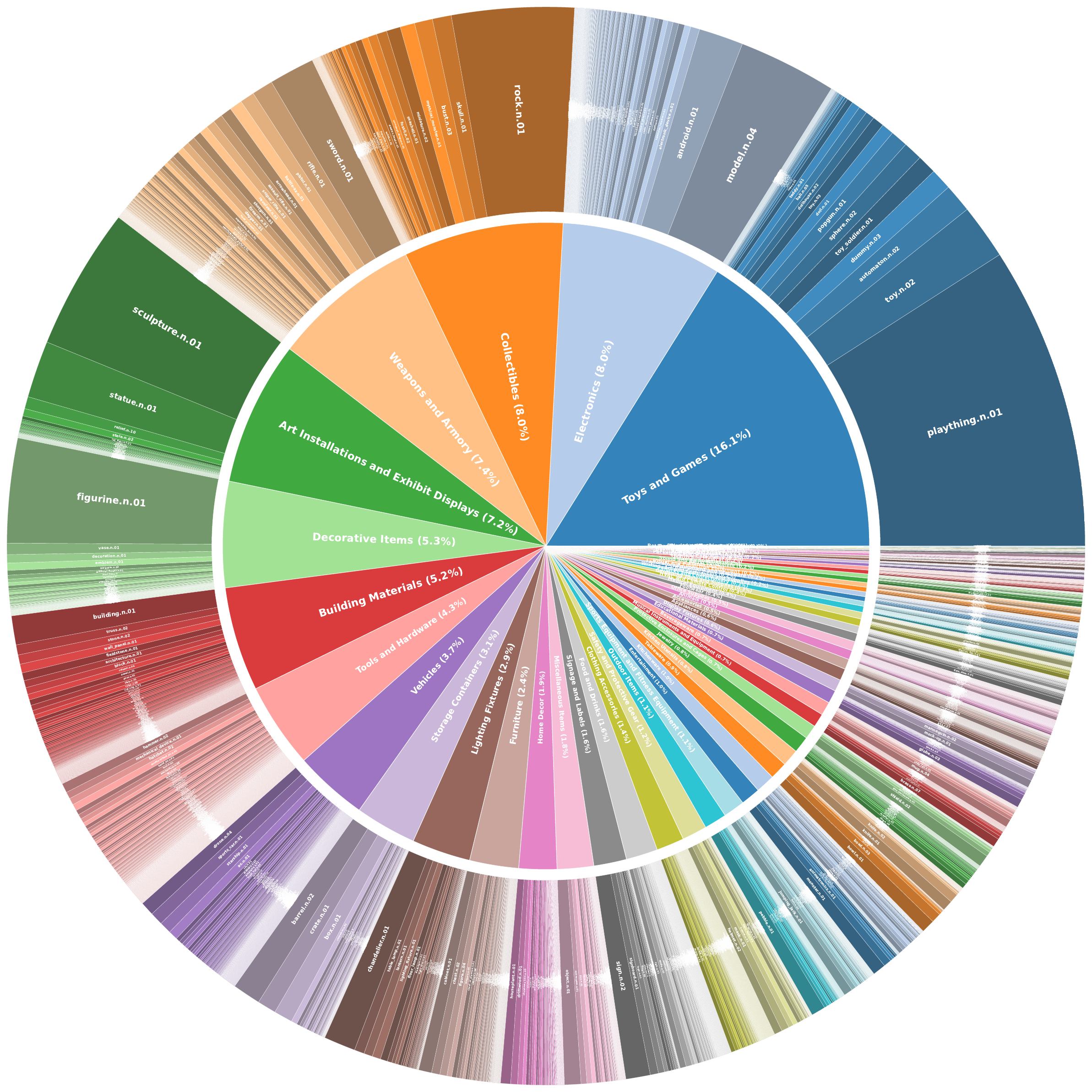}\\
    $\sim$110k \holodecklong scene specifications & $\sim$129k Objaverse objects
    \end{tabular}
    }
    \caption{Scene specification distribution (left) --with generic and concrete scene types and room types-- used to generate \holodecklong scenes. Between one and ten rooms of mostly scene-specific room types (here shown aggregated per generic scene type) are chosen to prompt LLMs.
    Right, distribution of WordNet synsets grouped by higher-level object categories in our curated subset of Objaverse.}
    \label{fig:holodeck_objaverse_distribution}
\end{figure*}

We provide models from both AI2‑THOR and Objaverse.
We extracted and converted objects from AI2‑THOR and migrated them into a file format compatible with MuJoCo as well as other simulators such as ManiSkill and IsaacSim. In total, we converted 1.6k rigid object instances across 134 categories into MuJoCo. In addition, there are 22 categories—including doors, refrigerators, and dressers—that were made articulable by annotating joint information, including joint type (slide or hinge), joint axis, joint position, and joint range.

To ensure physical realism for robot manipulation tasks, we performed several validation steps. For rigid objects, we verified that mass and density values were realistic by comparing simulated values with estimates annotated via large language models (LLMs), adjusting density as needed. For articulable objects, we created a teleoperation suite to manually tune the physical properties of joints and the density of movable parts. This was performed using a simulated Franka FR3 robot, which itself was tuned through system identification: real robot trajectories of picking and pushing cubes of known weights were collected, and simulation parameters were optimized to match the observed motions. These physics parameters were then applied to all objects during scene generation.

Collider meshes for all objects were generated using COACD \cite{wei2022coacd} , and we also annotated primitive colliders for all THOR objects. For physics stability, rigid objects with receptacles (e.g., tables, dressers) primarily use primitive colliders to avoid mesh-mesh contact issues. Manipulable objects require higher fidelity, so convex decomposition was used; however, for very small and thin objects, primitive colliders were employed to maintain simulation stability. Meshes were further processed and decimated to improve simulation performance.

In addition to AI2‑THOR objects, we converted a curated selection of Objaverse objects into the MJCF format for MuJoCo. This required a careful curation methodology to ensure quality and compatibility, which follows multiple filtering stages applied on top of an initial subset of 625k Objaverse objects pre-filtered with their original metadata, which are (1) converted to a format compatible with AI2-THOR \cite{kolve2017ai2thor,ai22024objathor} and (2) annotated with VLM-generated \cite{hurst2024gpt} descriptions, estimated mass, canonical poses, pickable/receptacle properties, matching Wordnet synsets \cite{mccrae2019wordnet, mccrae2022wordnet}, and scale estimates obtained via prompting with renderings from different viewpoints.

The automatic process results in some malformed or unreliable annotations, which prompts us to perform a complementary GPT-4.1 annotation (LLM prompt listed in Fig.~\ref{fig:reannotation-prompt}) to determine (1) the type and number of object instances in each model file, (2) the presence of props or other excessive geometry not part of the main object in the model file, (3) missing geometry, which typically occurs with scans of real-world objects, (4) a texture quality score on a 0-9 scale, and (5) whether the model contains receptacle parts. While we seek extraction of very specific information to help filter objects towards usability for embodied agent training, we note other efforts in re-annotating Objaverse like \cite{lin2025objaverseplusplus}.

To provide a subset of objects suitable for LLM-backed scene generation, we sequentially apply filtering stages to ensure (1) metadata completeness and synset validity, (2) presence of a single object in the model file, (3) statistical scale inliers using a restrictive Tukey rule on the IQR of the log-scale, (4) sufficient texture quality with annotated score 4 or higher, (5) cross–renderer fidelity with CLIP \cite{ clip, ilharco2021openclip, Cherti_2023_CVPR} similarity score 0.6 or higher, (6) processed model file size with compressed geometry and colliders less than 1.5 MB, (7) agreement on the receptacle property in both annotations, (8) watertight colliders with at least 80\% of the horizontal surface for receptacle objects, and (9) final removal of singleton representatives of synsets.
The resulting distribution of curated objects, corresponding to almost 2.8k different WordNet synsets, is illustrated by~\cref{fig:holodeck_objaverse_distribution} (right).
The count of nearly 129k curated objects, which more than doubles the available amount in prior work on LLM-backed scene generation \cite{yang2024holodeck}, is finally divided into train (80\%), validation (10\%), and test (10\%) splits. 

We further extend the filtering stages to provide compatibility with procedural house generation via ProcTHOR \cite{deitke2022procthor} relying on annotated placement options (prompt used with GPT-4o listed  in Fig.~\ref{fig:placement-prompt}). Rather than generating new placement options for all synsets, we rely on the already diverse set of THOR object categories and map each synset to the most relevant THOR category, if any, using scale and semantic similarity-based heuristics. In more detail, for each candidate pairing, we compute a weighted dissimilarity score based on room-type compatibility, feasible placement locations, object scale ranges, boolean affordances, and calibrated WordNet semantic similarity. Candidate matches are filtered using hard constraints on room and scale compatibility, followed by a lightweight reranking stage that emphasizes semantic alignment. Each synset is assigned to the lowest-cost THOR category below a fixed threshold, yielding a robust many-to-one alignment suitable for downstream scene generation. We then let the procedural engine sample objects from any synset associated with the currently selected THOR category. Using these heuristic placement reference assignments, the additional filtering stages ensure that objects (1) have synsets with valid placement references, (2) have annotated pickable/receptacle properties compatible with those of the placement reference, (3) have bounding box scales suitable for placement in houses, (4) are sufficiently shallow if wall-placeable, (5) have synsets that are not hyponyms of weapons, and (6) have synsets present in the train split. The result of applying these additional filters is a curated subset of 92.5k objects, corresponding to $\sim$2k synsets. 

For all objects, we provide an extensive metadata collection that includes the physical parameters of scale and mass as well as semantic information: name, category, and synsets. To enable the easy use of these models in robotics simulations, we also provide convex meshes and grasps. In addition to this, we make sure that these objects are defined in a canonical coordinate system to make them easily loadable into scenes.

\section{\holodecklong Generation Details}
\label{sec:holodeck_details}

\begin{figure*}
\centering
\begin{tabular}{ccc}
Original Holodeck \cite{yang2024holodeck} prompt & No edge placement emphasis & New prompt with grid layout \\
\midrule
\footnotesize{\parbox{0.3\linewidth}{\texttt{... edge: at the edge of the room, close to the wall, most of the objects are placed here... I prefer objects to be placed at the edge (the most important constraint) of the room if possible, which makes the room look more spacious...}}} &
\footnotesize\parbox{0.3\linewidth}{\texttt{... edge: at the edge of the room, close to the wall\\\textcolor{red}{- , most of the objects are placed here... I prefer objects to be placed at the edge (the most important constraint) of the room if possible, which makes the room look more spacious}...}} &
\footnotesize\parbox{0.3\linewidth}{\texttt{... \textcolor{darkgreen}{+ Using a grid constraint helps me create more realistic layouts when placing multiple objects in a repetitive pattern across a floor area that are meant to be away from room edges. Since grids are likely to occupy a considerable amount of space, it is convenient to place them as early as possible after the anchor object}...}} \\
\midrule
\includegraphics[width=0.2\linewidth]{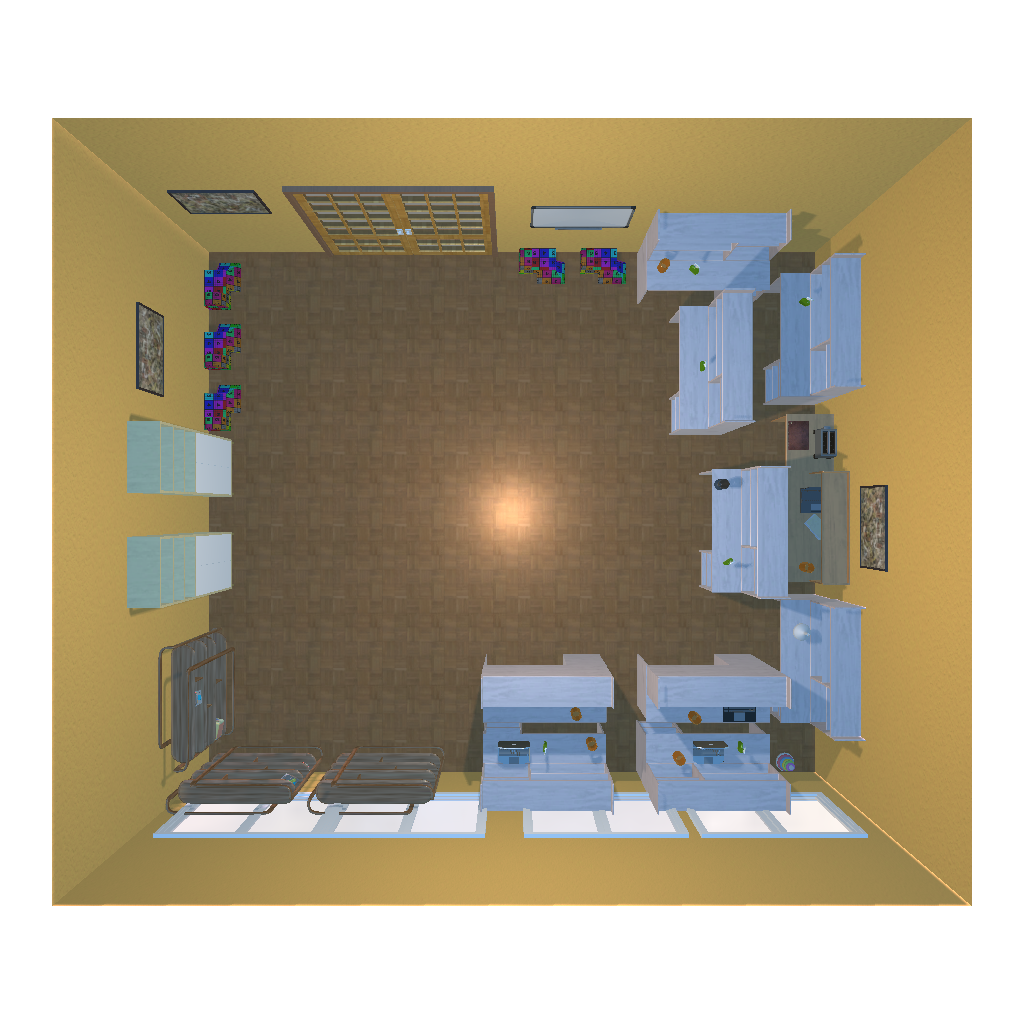} &
\includegraphics[width=0.2\linewidth]{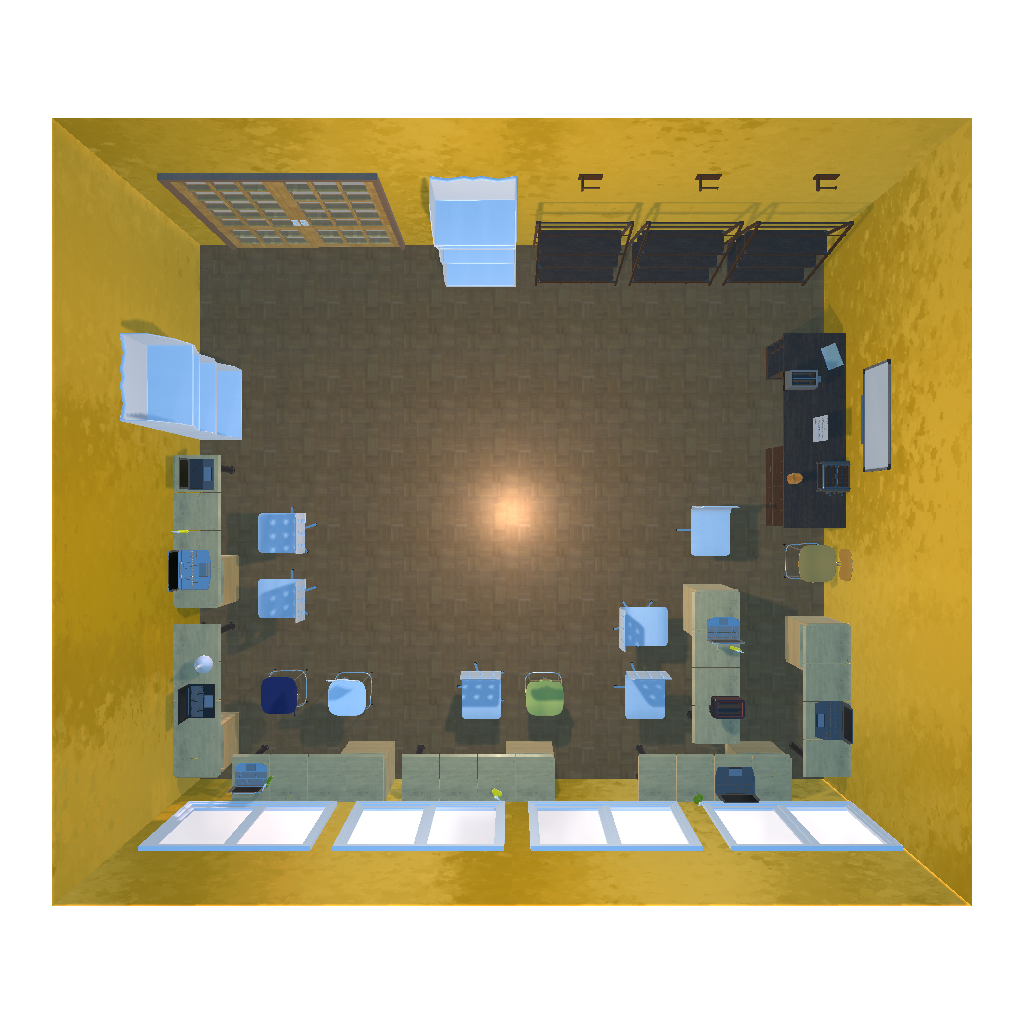} &
\includegraphics[width=0.2\linewidth]{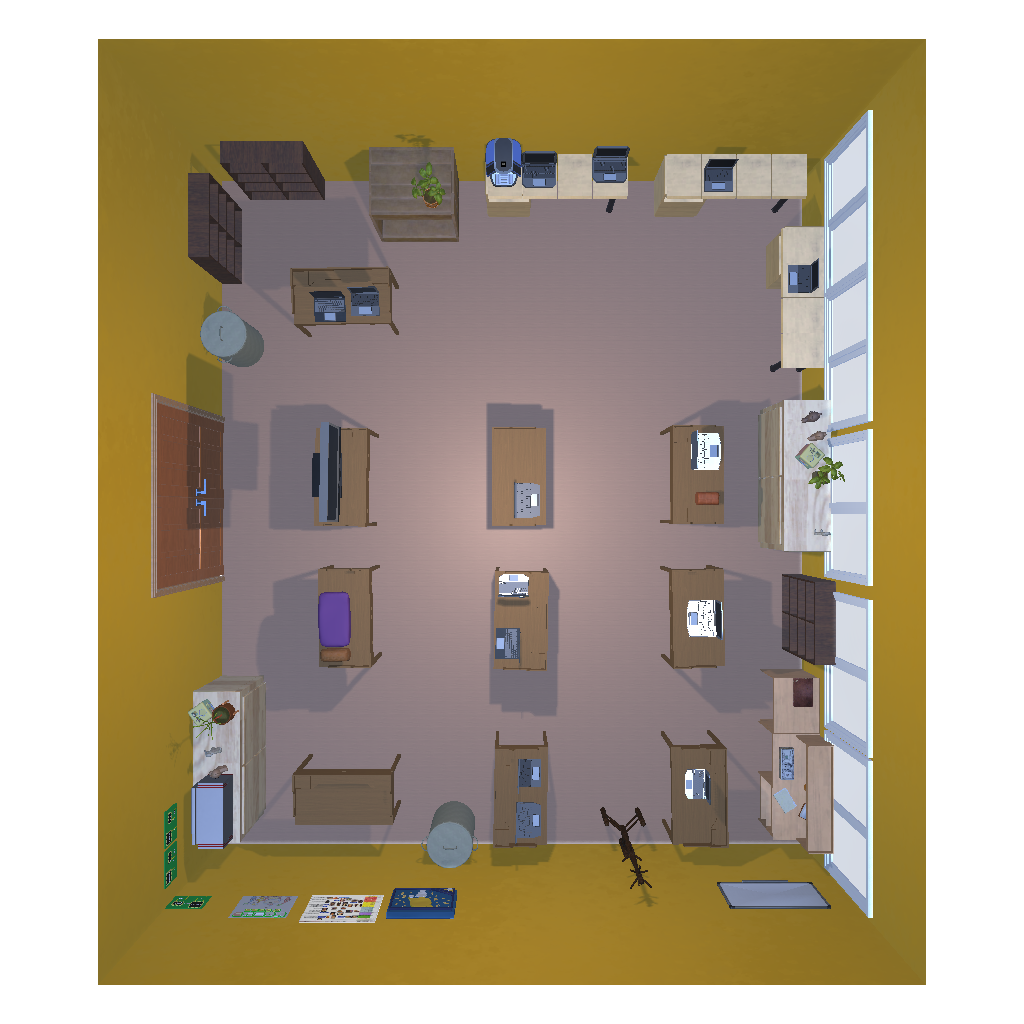} \\
\includegraphics[width=0.2\linewidth]{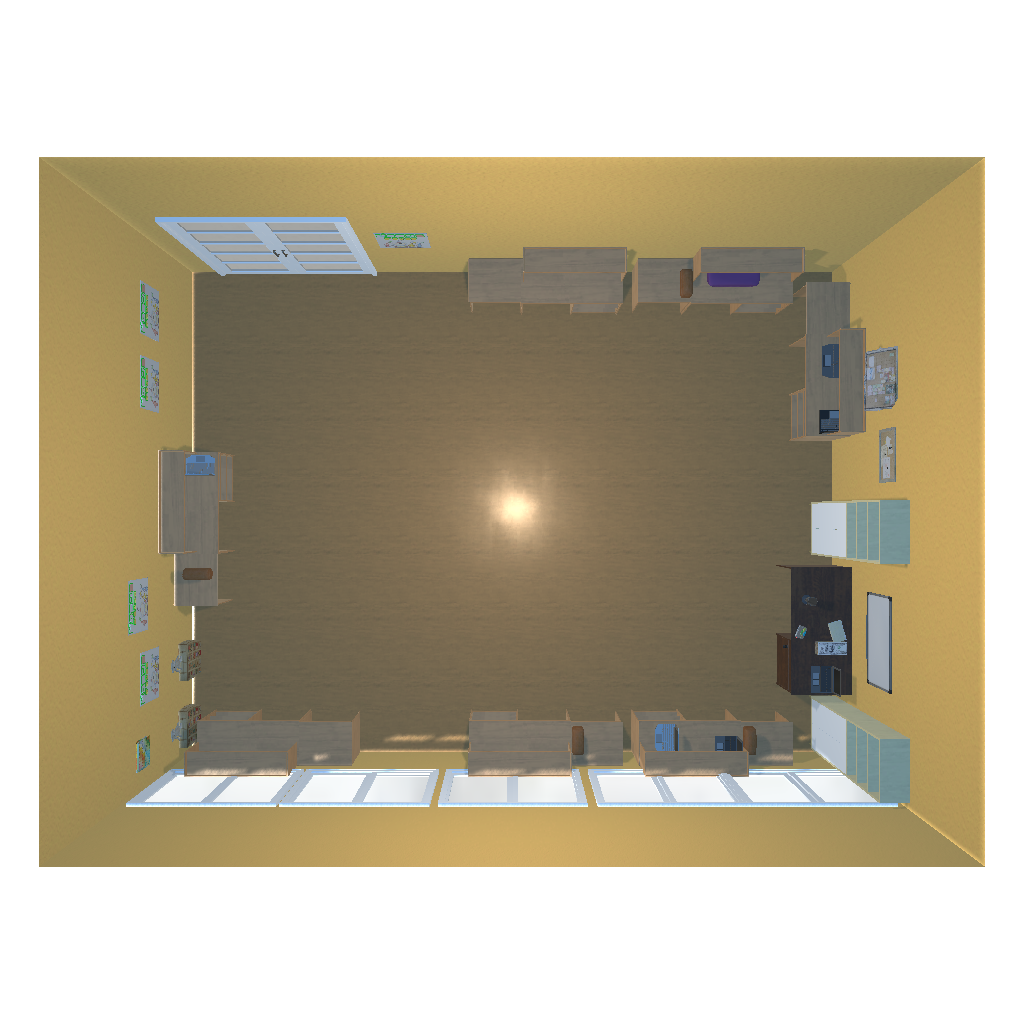} &
\includegraphics[width=0.2\linewidth]{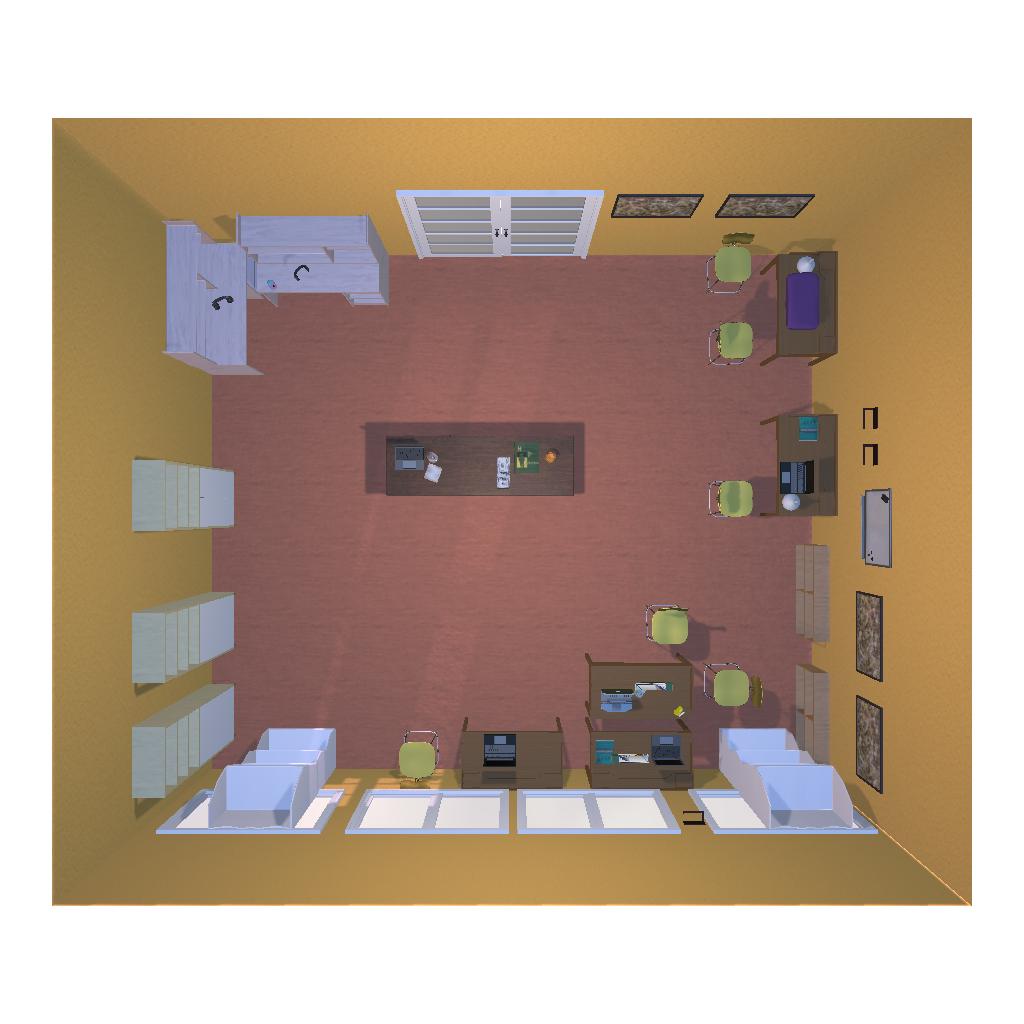} &
\includegraphics[width=0.2\linewidth]{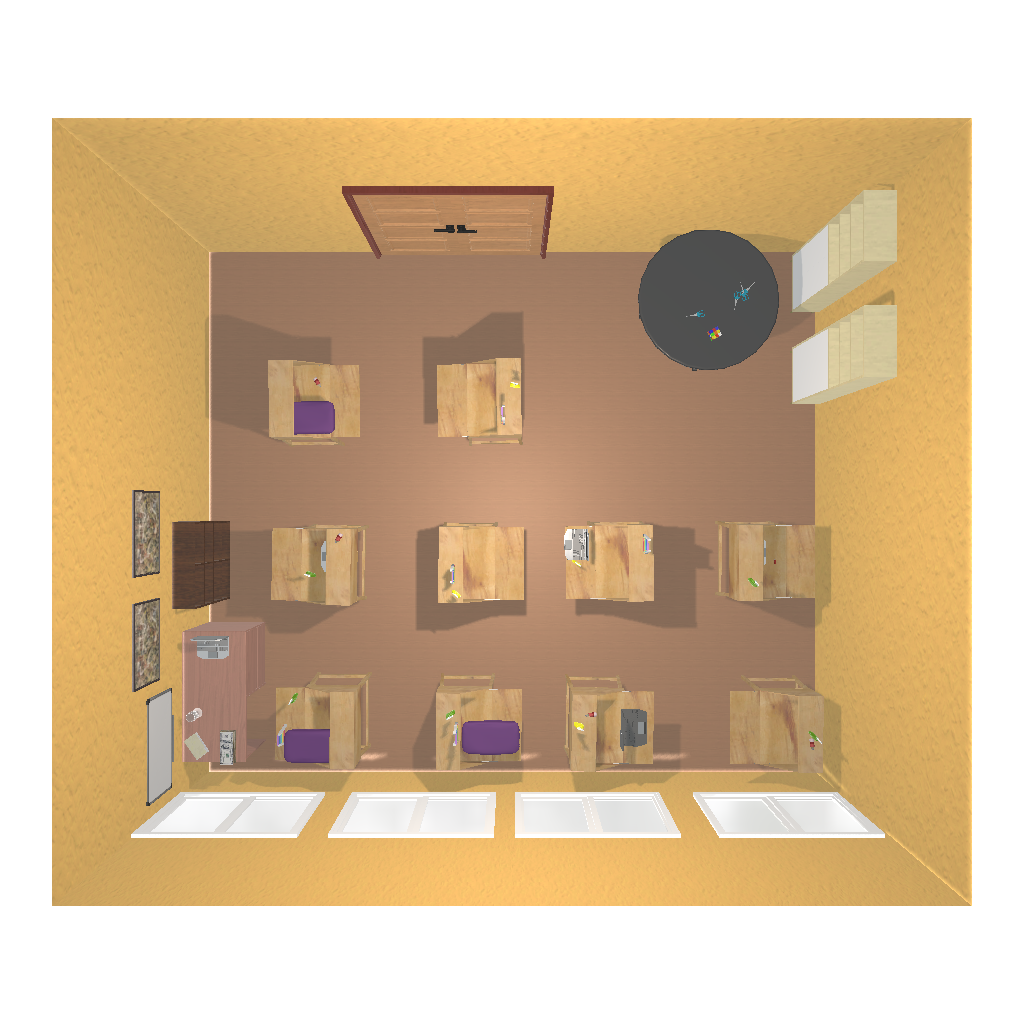} \\
\includegraphics[width=0.2\linewidth]{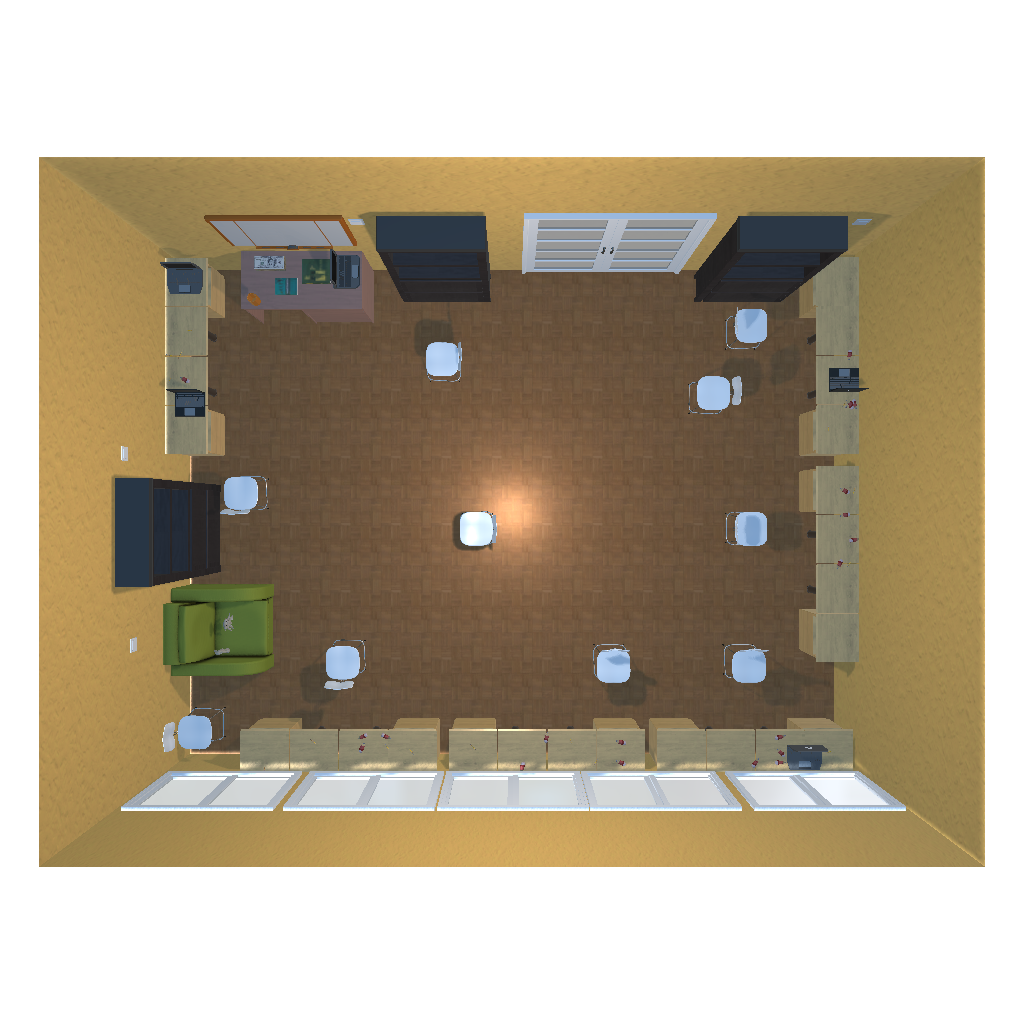} &
\includegraphics[width=0.2\linewidth]{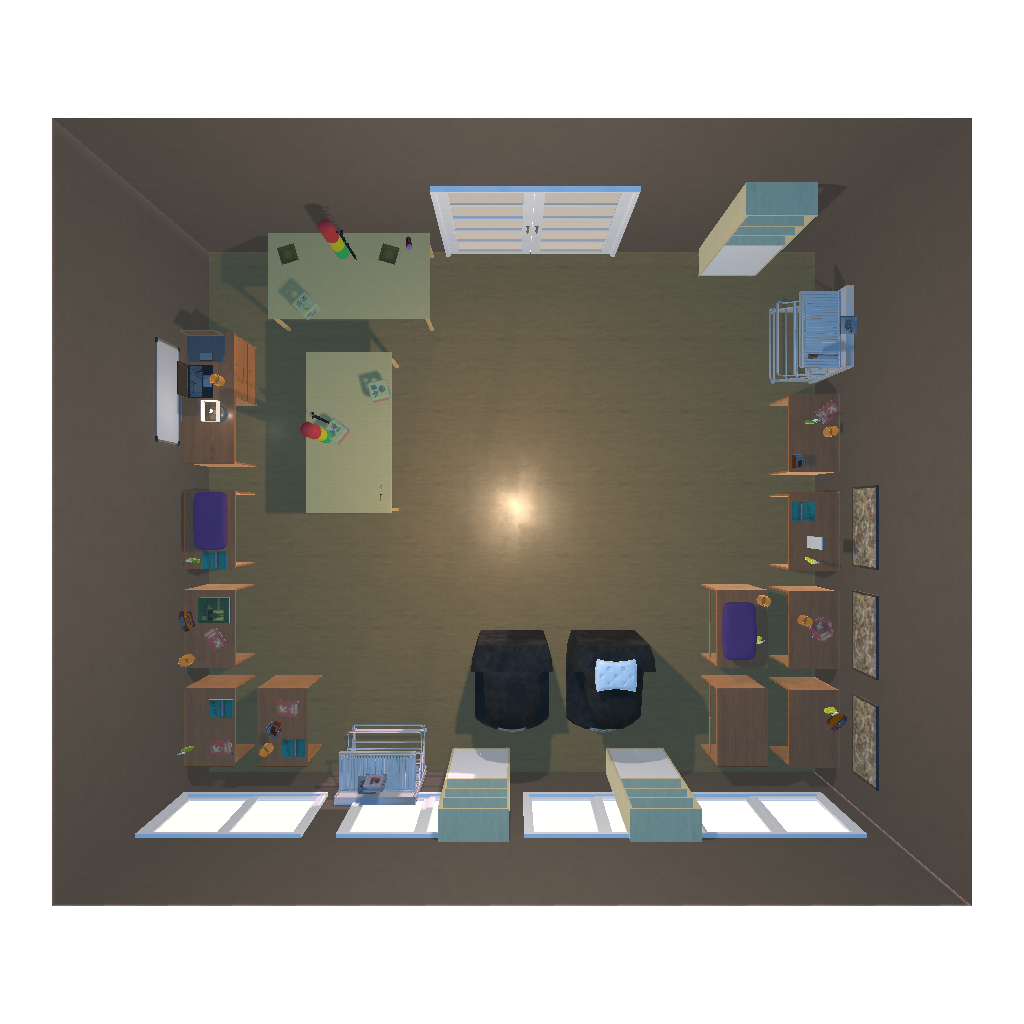} &
\includegraphics[width=0.2\linewidth]{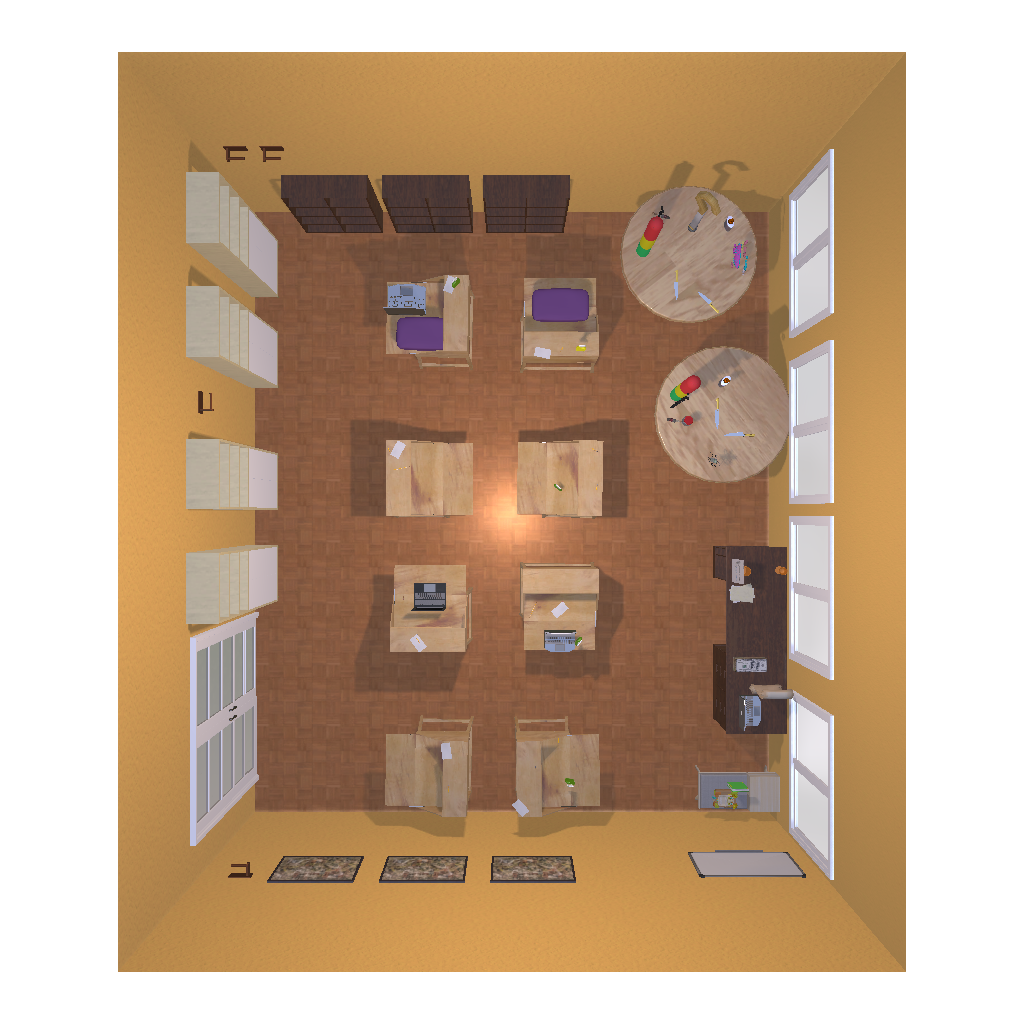} \\
\end{tabular}
\caption{\label{fig:grid-layout}Object placement versus inclusion of the `grid' layout constraint. In all samples, the textual scene specification in the LLM prompt is `\texttt{a primary school classroom}'. In the bottom row we additionally disable a bias term meant to improve door-to-door navigability in multi-room scenes. Text in red (green) indicates removal (addition).}
\end{figure*}

\paragraph{Grid layout placement option}
As mentioned in Sec.~\ref{sec:scene_datasets}, we extend the original DFS-based floor object solver in Holodeck \cite{yang2024holodeck} with a `grid' global constraint that enables structured batch placement of multiple objects suitable for several non-residential \holodeck scene types. Objects to be placed with a grid constraint are handled jointly: the solver enumerates feasible grid shapes (rows $\times$ columns) that can accommodate an iteratively decreasing object set --starting with the original object count requested by the LLM--, prioritizing compact layouts. For each candidate grid shape, the solver attempts to place the entire grid footprint within the current room and greedily assigns objects to grid cells by selecting positions and rotations that maximize satisfaction of remaining (non-grid) constraints regarding already placed objects. Grid configurations are scored by aggregating per-object constraint satisfaction, including an additional soft bias discouraging obstruction of door-to-door circulation. The highest-scoring grid placement is retained before resuming DFS over the remaining objects. We optionally apply a small positional and rotational jitter to grid-constrained objects, retaining only collision-free perturbations. Fig.~\ref{fig:grid-layout} illustrates how the new grid layout modality can simplify the task of placing uniformly spaced objects.

\begin{figure*}
\centering
\begin{tabular}{cccc}
\midrule
\footnotesize{\parbox{0.2\linewidth}{\texttt{a waiting room}}} & \footnotesize{\parbox{0.2\linewidth}{\texttt{the favorite waiting room of a person who identifies as `A vintage vinyl record collector who is challenged to keep their growing collection in check`}}} & \footnotesize{\parbox{0.2\linewidth}{\texttt{the favorite waiting room of a person that identifies as `A curator specializing in health and science exhibits, constantly seeking the epidemiologist's input to ensure accuracy and educational value`}}} & \footnotesize{\parbox{0.2\linewidth}{\texttt{the favorite waiting room of a person who identifies as `A former PBA basketball player who still holds a grudge against the alumni for leaving the team`}}} \\
\midrule
\includegraphics[width=0.22\linewidth]{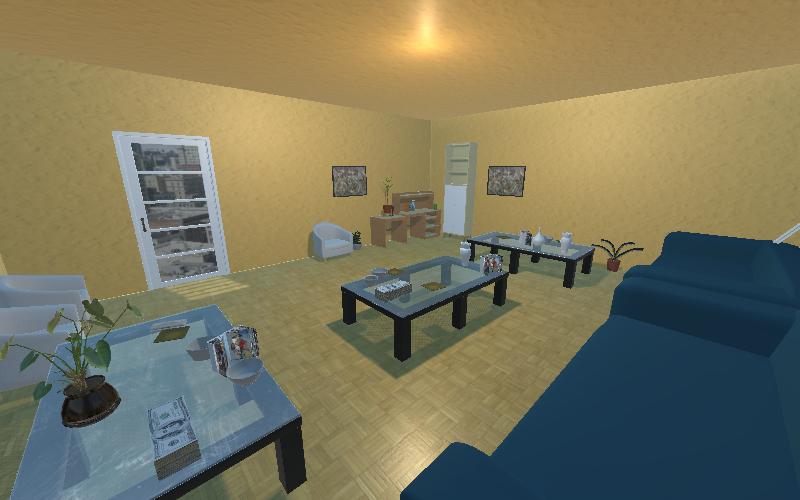} &
\includegraphics[width=0.22\linewidth]{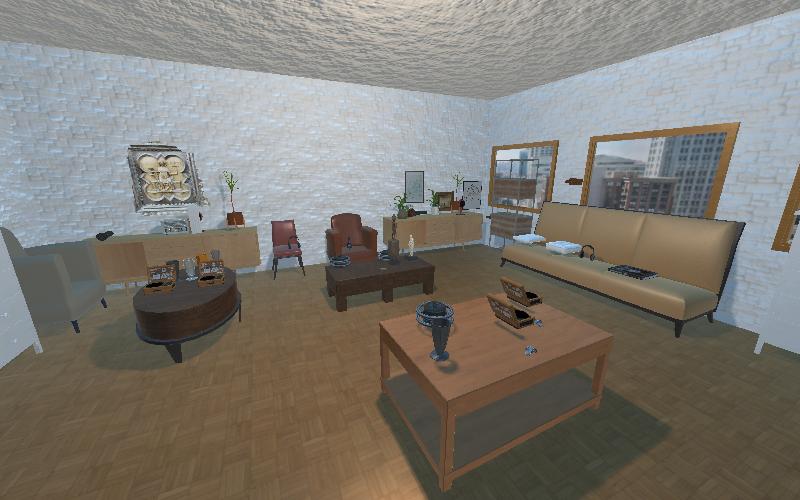} &
\includegraphics[width=0.22\linewidth]{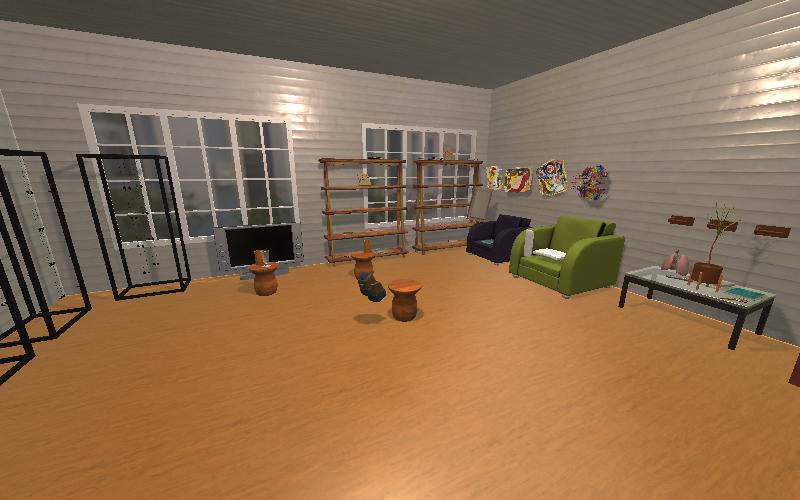} &
\includegraphics[width=0.22\linewidth]{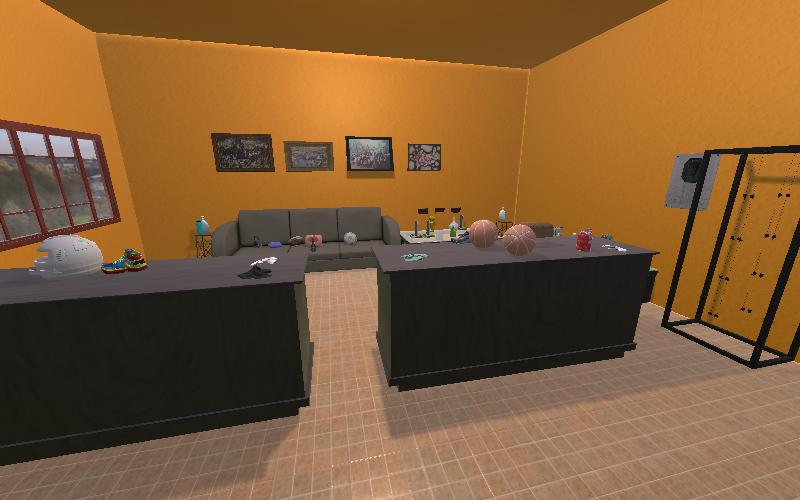} \\

\end{tabular}
\caption{\label{fig:persona} Complementing scene specifications with persona descriptions produces notable stylistic changes and inclusion of a wider variety of object types in LLM-generated scenes.}
\end{figure*}

\paragraph{Style accents with persona descriptions}
Persona descriptions conveying stylistic information were filtered from the full set in PersonaHub \cite{ge2024scaling} via GPT-4.1-mini prompted as shown in Fig.~\ref{fig:persona-selection prompt}. The effects of adding samples from this filtered set to a scene specification via the simple binding structure used to generate the $\sim$110k scenes in \holodeck are illustrated by Fig.~\ref{fig:persona}.

\begin{figure}[!b]
    \centering
    \footnotesize\fontfamily{zi4}\selectfont
    \begin{tcolorbox}[
      width=\linewidth,
      colback=gray!5,
      boxrule=1pt,
      sharp corners,
    ]
SYSTEM PROMPT:
\\\\
You are an expert in 3D model analysis. Your main task is to identify 3D models that contain more than a single object (e.g., kitchens with cabinets and sinks, dining tables with chairs, or assortments of objects of some or multiple categories, to name a few examples). You also decide whether the 3D model seems to have missing geometry, e.g. resulting from an incomplete scan, or excessive geometry, e.g., resulting from a scan that also reconstructs part of the supporting surface or background, or due to additional props added as decoration for rendering. Among others, a model without excessive geometry should not make the main object appear as being mounted or placed on any type of surface or structure. Finally, you also decide whether the model's object contains surfaces that can be used as receptacles of smaller object types (like seats of a chair or sofa, or top of a table, among others), and whether the model seems realistically textured for an object of the given category.
\\\\
The provided data is a collage of renders of the 3D model from different perspectives, showing the front, left, back, and right sides. The expected background should be a flat white color, with everything else being part of the model.
\\\\
Your decisions should be structured as a JSON dict with the following entries:
\begin{itemize}
\item[-]`object types': List[str], with each entry corresponding to an object type present in the model, e.g., ['shoe'] for a model of a pair of shoes;
\item[-]`object instances': Dict[str, int], with an approximate count of instances of each type, e.g., {'book': 3} for a model of a pile of 3 books);
\item[-]`main object type': str, type of either largest object, or the most meaningful one, e.g., 'plate' for a model of a plate with a fork on top;
\item[-]`supporting structures': List[str], typically empty, with names of visible structures supporting the main object, e.g. fragments (or panels) of tables, walls, floors, rugs, etc.;
\item[-]`excessive geometry': str, typically `No excessive geometry', briefly describing the presence of supporting or background surfaces possibly added to the model as decoration for rendering or due to poor segmentation of a reconstructed or scanned model;
\item[-]`has excessive geometry': bool, providing a binary response from the analysis under `excessive geometry';
\item[-]`detachable part types': List[str], typically empty, including easily detachable parts of the main object that should not be considered additional objects as they appear in place, e.g., [`light bulb', `lamp shade'] for a model of a table lamp with those parts visible and mounted at their proper places on the main object;
\item[-]`is single object': bool, considering all previous responses, whether the model only contains one object (no additional objects, other than possibly detachable parts that appear properly installed on the object) and does not include excessive geometry of any kind;
\end{itemize}
...
    \end{tcolorbox}
    \caption{\label{fig:reannotation-prompt}System prompt used with GPT-4.1 to generate alternative annotation for Objaverse object assets filtering in batch mode (continues in Fig.\ref{fig:reannotation-prompt-continued}).}
\end{figure}

\begin{figure}[!b]
    \centering
    \footnotesize\fontfamily{zi4}\selectfont
    \begin{tcolorbox}[
      width=\linewidth,
      colback=gray!5,
      boxrule=1pt,
      sharp corners,
    ]
...
\begin{itemize}
\item[-]`missing geometry': str, typically `No missing geometry', briefly describing where the model appears to have missing geometry, e.g., no geometry around the back side, resulting in empty render or some distorted view of the front one for that perspective. Do not take into consideration whether the model appears to be lacking textures;
\item[-]`has missing geometry': bool, providing a binary response from the analysis under `missing geometry';
\item[-]`receptacles': List[str], generally empty for small objects, including names of the main object's surfaces that can be used as receptacles for smaller objects, as their normals are oriented upward and have sufficient area with no/low curvature, e.g. [`top of mattress'] for a model of a bed with an installed mattress.
\item[-]`texture quality': int, range 0 (no texture, making the object in the model hard to identify) to 9 (detailed and realistic texture for the object type, making the model appear close to real).
\end{itemize}
An example response to a query with renders of a model depicting a bookshelf with about 10 books in red, blue, and white placed on a green tiled floor where the render from the back shows the same spines seen from the front could be:
\\\\
<EXAMPLE JSON OMITTED>
\\\\
Feel free to briefly reason before providing you response as a JSON parseable dict, which must include clear and concise values for all the required entries without requesting additional input.
    \end{tcolorbox}
    \caption{\label{fig:reannotation-prompt-continued}System prompt used with GPT-4.1 to generate alternative annotation for Objaverse object assets filtering in batch mode (continues from Fig.\ref{fig:reannotation-prompt}).}
\end{figure}

\begin{figure}[!b]
    \centering
    \footnotesize\fontfamily{zi4}\selectfont
    \begin{tcolorbox}[
      width=\linewidth,
      colback=gray!5,
      boxrule=1pt,
      sharp corners,
    ]
We need to generate placement options for synsets used to label objects. For each of these synsets, we have associated specific types, text descriptions of some objects, and scale ranges.
\\\\
Write the placement options in JSON format, and do not add any additional comments. The structure is a mapping from each synset to a nested mapping with Boolean entries, indicating whether each property seems likely for objects of the given synset:
\\
\{
\begin{itemize}
    \item[] `hasMultipleObjects': refers to a composition or set of several objects,
    \item[] `isScene': refers to a scene or a set of objects (like an assortment) rather than a single object,
    \item[] `roomTypes': \{
    \begin{itemize}
        \item[] `inKitchens': appears in a kitchen,
        \item[] `inLivingRooms': appears in a living room, media room, or dining room,
        \item[] `inBedrooms': appears in a bedroom, office, or playroom,
        \item[] `inBathrooms': appears in a bathroom, restroom, or laundry room,
        \item[] `inOthers': appears in other room types like garages, balconies, etc.
    \end{itemize}
    \item[] \},
    \item[] `feasibleLocations': \{
        \item[] `onFloorInCorner': placed directly on the floor, in a corner,
        \item[] `onFloorInMiddle': placed directly on the floor, anywhere away from walls,
        \item[] `onFloorOnEdge': placed directly on the floor and in contact with a wall,
        \item[] `onWall': placed on a wall,
        \item[] `fromCeiling': hangs from the ceiling
    \item[] \},
    \item[] `isPickupable': allows being picked up with a single hand,
    \item[] `isKinematic': has an effective fixed pose, as in Unity's kinematic bodies,
    \item[] `multiplePerRoom': appears multiple times in the same room
\end{itemize}
\}
\\\\
Please set at least one of the `roomTypes' (try to use `ìnOthers' sparingly) unless the synset seems to refer to a scene rather than a single object as signaled in `ìsScene'. Also note that enabling any of the `feasibleLocations' options (floor, wall, or ceiling) prevents the objects os the synset from being placed on top of other structures or furniture.
\\\\
The synsets (along with associated specific types, sample descriptions, and scale ranges in cm) to annotate are:
\\\\
<OMITTED BATCH OF INPUT DATA>
    \end{tcolorbox}
    \caption{\label{fig:placement-prompt}Prompt used with GPT-4o to determine placement options for a given batch of synsets.}
\end{figure}

\begin{figure}[!b]
    \centering
    \footnotesize\fontfamily{zi4}\selectfont
    \begin{tcolorbox}[
      width=\linewidth,
      colback=gray!5,
      boxrule=1pt,
      sharp corners,
    ]
SYSTEM PROMPT:
\\\\
You are an expert in sociology, psychology, and interior design counseling. Given a list of one-line personal identity statements, your task is to identify which individuals are likely to find certain indoor scenes visually memorable due to the presence of large, interest-relevant objects. These objects should stand out even in visually cluttered environments.
\\\\
Typically, only about 25\% of the identities will be sufficiently specific or interest-oriented to warrant this kind of visual sensitivity. Avoid selecting identities that are too abstract, ethnicity-related, enterprise-centric, or related solely to virtual environments, as these are unlikely to correspond to specific, visually memorable physical elements.
\\\\
Return your output as a JSON-parseable dict mapping indices corresponding to identity statements that are useful for informing visually impactful interior design styles to a string describing visual objects, decoration items, pieces of furniture, etc., that would instantly draw the attention of each chosen personality in some indoor scene.
    \end{tcolorbox}
    \caption{\label{fig:persona-selection prompt}System prompt used with GPT-4.1-mini to select valid persona descriptions in batch mode.}
\end{figure}

\newpage
\begin{figure*}[h]
    \centering

    \begin{subfigure}{1.0\linewidth}
        \centering
        \includegraphics[width=\linewidth]{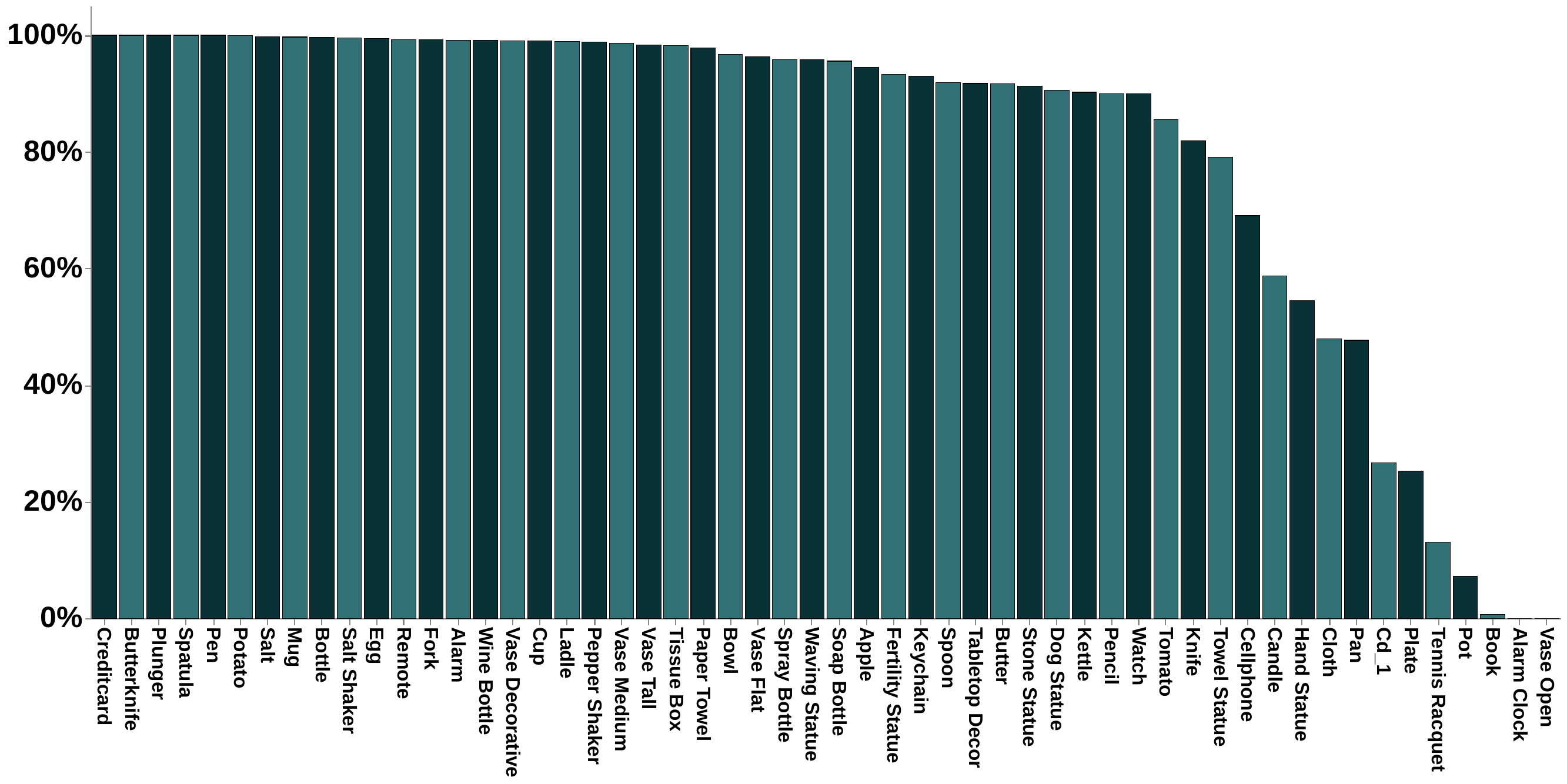}
        \caption{Average grasp success rates by object category in a random sample of MolmoSpaces houses.}
        \label{fig:pickup_thor_categories}
    \end{subfigure}

    \vspace{0.5cm}

    \begin{subfigure}{1.0\linewidth}
        \centering
        \includegraphics[width=\linewidth]{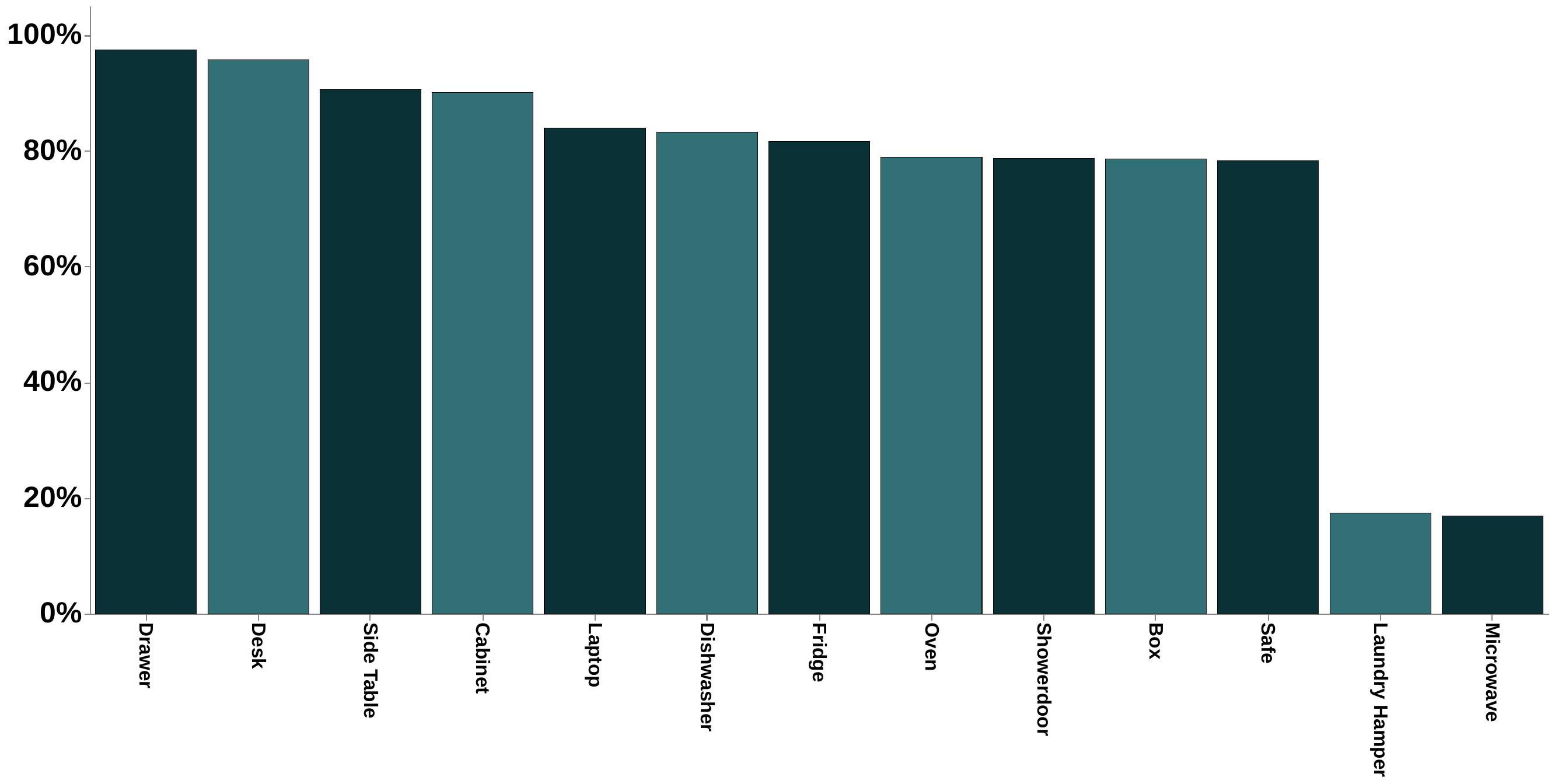}
        \caption{Average grasp success rates by category for articulated objects in a random sample of MolmoSpaces houses.}
        \label{fig:articulate_thor_categories}
    \end{subfigure}

    \caption{Average grasp success rates by category for all objects tested in a random sample of scenes from MSCrafted and MSProcObja MolmoSpaces-Scene dataset.}
    \label{fig:grasp_tests_thor_categories}

\end{figure*}

\begin{table*}[t]
\centering
\caption{Comparison of large-scale grasp datasets. MolmoSpaces provides the highest number of object instances.}
\begin{tabular}{lcccc}
\toprule
\textbf{Dataset} & \textbf{Year} & \textbf{\# Objects} & \textbf{\# Grasps} \\
\midrule
HO-3D & 2020 & 10 & 78,000 \\
EGAD & 2020 & 2,331 & 233,000 \\
DDG & 2020 & 500 & 50,000 \\
DexYCB & 2021 & 20 & 582,000 \\
Acronym & 2021 & 8,872 & 17,000,000 \\
UniGrasp & 2020 & 1,000 & 2,000,000 \\
DexGraspNet & 2023 & 5,355 & 1,300,000 \\
Fast-Grasp’D & 2023 & 2,350 & 1,000,000 \\
GenDexGrasp & 2023 & 58 & 436,000 \\
MultiGripperGrasp & 2024 & 345 & 30,400,000 \\
GraspGen & 2025 & 8,515 & \textbf{53,100,000} \\
\midrule
\textbf{MolmoSpaces-Grasps (Ours)} & \textbf{2026} & \textbf{48,675} & 42,000,000 \\
\bottomrule
\end{tabular}
\label{tab:grasp-comparison}
\end{table*}


\end{document}